\documentclass[paper]{ieice}
\usepackage[pdftex]{graphicx,xcolor}
\usepackage[fleqn]{amsmath}
\usepackage{newtxtext}
\usepackage[varg]{newtxmath}
\usepackage{multirow}
\usepackage{url}
\usepackage[subrefformat=parens]{subcaption}
\newtheorem{defi}{Definition}

\field{A}
\title{\title{Unified Likelihood Ratio Estimation for High- to Zero-frequency $N$-grams}}
\authorlist{
 \authorentry[kikuchi@nitech.ac.jp]{Masato KIKUCHI}{m}{Nagoya}\MembershipNumber{}
 \authorentry{Kento KAWAKAMI}{n}{LINE}\MembershipNumber{}
 \authorentry{Kazuho WATANABE}{m}{Toyohashi}\MembershipNumber{}
 \authorentry{Mitsuo YOSHIDA}{n}{Toyohashi}\MembershipNumber{}
 \authorentry{Kyoji UMEMURA}{m}{Toyohashi}\MembershipNumber{}
}
\affiliate[Nagoya]{The author is with Department of Computer Science, Graduate School of Engineering, Nagoya Institute of Technology, Nagoya-shi, 466-8555 Japan.
}
\affiliate[Toyohashi]{The authors are with Department of Computer Science and Engineering, Toyohashi University of Technology,
 1-1 Hibarigaoka, Tempaku-cho, Toyohashi 441-8580, Japan.
}
\affiliate[LINE]{The author is with LINE Corporation, 4-1-6 Shinjuku, Shinjuku-ku,Tokyo160-0004, Japan.
}

\received{2015}{1}{1}
\revised{2015}{1}{1}

\begin{document}
\maketitle
\begin{summary}
Likelihood ratios (LRs), which are commonly used for probabilistic data processing, are often estimated based on the frequency counts of individual elements obtained from samples.
In natural language processing, an element can be a continuous sequence of $N$ items, called an $N$-gram, in which each item is a word, letter, etc.
In this paper, we attempt to estimate LRs based on $N$-gram frequency information.
A naive estimation approach that uses only $N$-gram frequencies is sensitive to low-frequency (rare) $N$-grams and not applicable to zero-frequency (unobserved) $N$-grams; these are known as the low- and zero-frequency problems, respectively.
To address these problems, we propose a method for decomposing $N$-grams into item units and then applying their frequencies along with the original $N$-gram frequencies.
Our method can obtain the estimates of unobserved $N$-grams by using the unit frequencies.
Although using only unit frequencies ignores dependencies between items, our method takes advantage of the fact that certain items often co-occur in practice and therefore maintains their dependencies by using the relevant $N$-gram frequencies.
We also introduce a regularization to achieve robust estimation for rare $N$-grams.
Our experimental results demonstrate that our method is effective at solving both problems and can effectively control dependencies.
\end{summary}

\section{Introduction}\label{sec:introduction}

Likelihood ratios (LRs)---statistical measures defined as ratios of likelihood functions---are used in numerous tasks, including relation extraction~\cite{Jo:07,Huang:14} and data analysis~\cite{Ravichandran:15,Tran:19}.
In practice, LRs are estimated using samples drawn from populations.
In natural language processing (NLP), observed frequencies of discrete elements, e.g., word frequencies observed from language resources, are often used for estimation~\cite{Dunning:93,Manning:99}.
In general, however, there will be a limited number of observed elements, and many of them will be infrequent.
In this paper, we focus on the problems that the well-used estimation framework encounters in the presence of low- or zero-frequency elements.

To clarify our motivation, we suppose that the prediction task for the left context of named entities (NEs), which are proper nouns (e.g., person names and location names), and calculate the following likelihood ratio to predict the context:
\begin{align*}
r(w) &= \frac{p_{\rm nu}(w)}{p_{\rm de}(w)},
\end{align*}
where $w$ is an element sampled from a probability distribution with density $p_*(w)$, $* \in \{\rm de, nu\}$.
Let $w=a_1a_2 \cdots a_{N}$ be a continuous sequence of $N$ items (called an $N$-gram) in which $a_k$ is an item such as a word or a letter at the $k$-th position.
$p_{\rm de}(w)$ is the probability density where $w$ occurs at any position in the corpus, and $p_{\rm nu}(w)$ is the probability density where $w$ occurs on the left side of the NEs.
That is, if $w$ is the left context of the NEs, $r(w)$ is high. Otherwise, $r(w)$ is low.
A naive estimation approach would be to first obtain the maximum likelihood estimate (MLE) $\widehat{p}_*(w)$ of each distribution and then take the ratio of MLEs\footnote{Strictly speaking, $\widehat{p}_*(w)$ indicates the MLE for the occurrence probability of $w$ when the occurrence is modeled using a multinomial distribution.}:
\begin{align}
\label{eq:Indirect_MLE}
\widehat{p}_*(w) &= \frac{c_*(w)}{n_*}, \nonumber \\
r_{\rm MLE}(w) &= \frac{\widehat{p}_{\rm nu}(w)}{\widehat{p}_{\rm de}(w)},
\end{align}
where $c_*(w)$ is the frequency count of $w$ sampled from the distribution with density $p_*(w)$ and $n_*$ is the sum of $c_*(w)$ for all $w$.
In this task, $c_{\rm de}(w)$ is the frequency of $w$ across the entire corpus, and $c_{\rm nu}(w)$ is the frequency of $w$ on the left side of the NEs.
We call this type of approach ``indirect estimation'' because it involves probability estimation.
As an example, we examine the frequencies listed in Table~\ref{tab:Frequencies_example}.
\begin{table}[!t]
\caption{Examples of occurrence frequencies.}
\renewcommand{\arraystretch}{1.3}
\label{tab:Frequencies_example}
\centering
	\begin{tabular}{ c  r  r  r  r  r } \hline
		\multicolumn{1}{ c }{$N$-gram} & \multicolumn{4}{ c }{Frequency}  & \multirow{2}{*}{$r_{\rm MLE}(w)$} \\ \cline{2-5}
		$w$ & \multicolumn{1}{ c }{$n_{\rm de}$} & \multicolumn{1}{ c }{$c_{\rm de}(w)$} & \multicolumn{1}{ c }{$n_{\rm nu}$} & \multicolumn{1}{ c }{$c_{\rm nu}(w)$} & \\ \hline \hline
		$w_{\rm A}$ & $10^7$ & 5,000 & $10^4$ & 100 & 20 \\
		$w_{\rm B}$ & $10^7$ & 50 & $10^4$ & 1 & 20 \\
		$w_{\rm C}$ & $10^7$ & 50 & $10^4$ & 2 & 40 \\
		$w_{\rm D}$ & $10^7$ & 14 & $10^4$ & 0 & 0 \\ \hline
	\end{tabular}
\end{table}
The estimates $r_{\rm MLE}(w)$ for each $w$ are listed in the rightmost column of the table.
Although this approach is simple, it faces the following two problems:
\begin{description}
\item[1)] {\bf Zero-frequency problem.}
$r_{\rm MLE}(w)$ for zero-frequency (unobserved) elements cannot be estimated.
For example, $r_{\rm MLE}(w_{\rm D})$ is zero regardless of the other frequencies because $c_{\rm nu}(w_{\rm D})$ is zero.
Furthermore, if $c_{\rm de}(w)$ is zero, $r_{\rm MLE}(w)$ will be infinity owing to division by zero.
Owing to the nature of the language, the more items, $\{a_k\}_{k=1}^N$, that compose $w$ (i.e., the larger $N$), the less likely the same $w$ occurs in the language resource.
However, in pre-learning systems, such as those of speech recognition and machine translation, long $N$-grams are often given as input to the trained models in the form of text or speech.
Then, most are not contained in the training set.
Thus, it is necessary to treat them as unobserved elements.
As described, if we set the estimates of the statistical measures for all unobserved elements to zeros or infinities, the systems often do not work well.
Therefore, it is important to properly estimate the measures, even for the zero-frequency elements.
Additionally, a real-time processing system must handle such elements without expanding the training set.
\item[2)] {\bf Low-frequency problem.}
$r_{\rm MLE}(w)$ estimated for low-frequency (rare) elements can have very large estimation errors.
For example, the estimation uncertainty for $w_{\rm B}$ is larger than that for $w_{\rm A}$ because the frequencies of $w_{\rm B}$ are lower than those of $w_{\rm A}$; that is, $c_{\rm de}(w_{\rm A})>c_{\rm de}(w_{\rm B})$ and $c_{\rm nu}(w_{\rm A})>c_{\rm nu}(w_{\rm B})$.
However, $r_{\rm MLE}(w_{\rm B})$ is high and equal to $r_{\rm MLE}(w_{\rm A})$ (both 20).
In particular, $c_{\rm nu}(w_{\rm B})=1$ may be a coincidence.
Therefore, $r_{\rm MLE}(w)$ should be adjusted depending on the uncertainty.
Moreover, $r_{\rm MLE}(w)$ is sensitive to the number of observations;
$c_{\rm nu}(w_{\rm C})$ has only one more observation than $c_{\rm nu}(w_{\rm B})$, but $r_{\rm MLE}(w_{\rm C})$ is twice $r_{\rm MLE}(w_{\rm B})$ (40 and 20, respectively).
In such cases, $r_{\rm MLE}(w)$ could be more robust.
Most $N$-grams are infrequent, and, as $N$ increases, the infrequent $N$-grams tend to increase further.
Therefore, to handle them, it is essential to address the low-frequency problem.
\end{description}

A method for mitigating the low-frequency problem was proposed by ~\cite{Kikuchi:19}.
This method applies {\it unconstrained Least-Squares Importance Fitting} (uLSIF)~\cite{Kanamori:09}, which is used to estimate $r(w)$ without estimates of $p _*(w)$ by solving a least-squares minimization and is therefore called ``direct estimation.''
The resulting model in~\cite{Kikuchi:19} underestimates $r(w)$ properly by using a regularization parameter introduced in the minimization process, thereby making the estimator robust (see Sect.~\ref{sec:Kikuchi19}).
To the best of our knowledge, no other method has been found to mitigate the low-frequency problem in LR estimation.
Unfortunately, this method does not address the zero-frequency problem.
In fact, in cases in which many rare elements are observed, it is expected that many unobserved elements will also be present.
Thus, the low- and zero-frequency problems should both be addressed at the same time if possible.

This paper presents a method for mitigating both problems based on the following approaches: \\
{\bf Approach for Problem 1.}
One possible approach for estimating $r(w)$ is to ``itemize'' or decompose $w$ into item units of $\{a_k\}_{k=1}^N$ and then estimate $r_{\rm item}(w)$ using their frequencies.
Itemizing $w$ is commonly performed in probability estimation and is here applied to LR estimation.
If each $a_k$ is observed separately, $r_{\rm item}(w)$ can be estimated even for unobserved $w$.
For example, we suppose that the entity type is person name and that $w=``{\rm to\ Mr.}"$ ($a_1 = ``{\rm to}"$ and $a_2 = ``{\rm Mr.}"$).
If the two words $a_1$ and $a_2$ occur separately, we can estimate $r_{\rm item}(w)$, even if $w$ does not occur.
However, the standard estimation of $r_{\rm item}(w)$ assumes that all items occur independently, and the dependencies between items are ignored.
Here, we give another example.
We suppose that the entity type is location name, and that $w=``{\rm Hospital\ in}."$
In this example, the dependency of $a_1$ and $a_2$ is important.
This is because the preposition ``in" occurs in various contexts, and the noun ``Hospital" at the left of ``in" is the key to predicting whether $w$ is the left context of location names.
To utilize dependency, we introduce the term $t_{\rm d}(w)$, which is estimated using the frequencies of $w$.
In this manner, our method can incorporate dependency and successfully treat observed and unobserved $w$. \\
{\bf Approach for Problem 2.}
An important challenge is determining the term $t_{\rm d}(w)$.
For $w$ with $c_{\rm de}(w)=c_{\rm nu}(w)=0$, we estimate $r_{\rm item}(w)$ and then set $t_{\rm d}(w)$ to zero.
For rare $w$, we can use an approach similar to that used in~\cite{Kikuchi:19}.
Both $r_{\rm item}(w)$ and $t_{\rm d}(w)$ are estimated using regularization parameters, which can mitigate the low-frequency problem.
More importantly, these parameters are shown to also be capable of controlling the dependency between items.
We explain this advantage by reusing the examples used in the approach for problem 1.
If $w = ``{\rm to\ Mr.},"$ we can imagine that a person name will be on the right side of $a_2=``{\rm Mr.},"$ regardless of $a_1$.
Therefore, the required dependency between $a_1$ and $a_2$ is inferred to be weak.
In contrast, in the case of $w = ``{\rm Hospital\ in},"$ the sequence of two words is important for context prediction, and the required dependency is inferred to be strong.
In such examples, where the required dependencies are different, the proposed method adjusts the strength of the dependency and incorporates the appropriate dependency into the estimate.
Consequently, our method can obtain an informative estimate, even for rare $w$.

Experimental results demonstrate that our method is effective at addressing both problems and further enables the control of dependency between items, which leads to performance improvement.

\section{Related Work}\label{sec:Related_Work}

Naive LR estimation approaches face both the low- and the zero-frequency problems.
As a first step toward finding a mitigation approach, in this Sect. we review several methods that have been developed to address these problems for non-LR statistical measures.

Studies addressing the use of estimation uncertainty to mitigate the low-frequency problem include Johnson~\cite{Johnson:99}, who used confidence interval bounds for pointwise mutual information (PMI) estimation.
Such bounds can reflect the estimation variance because the interval varies with it.
Pantel and Ravichandran~\cite{Pantel:04} proposed a penalized estimator based on frequencies.
Because PMI includes LR in its definition, these two measures are closely related.
As a conditional probability estimation method, Rudin et al.~\cite{Rudin:13} proposed an approach involving obtaining the MLE and then adding a positive constant to the denominator.
As conditional probability can be defined as a probability ratio, it can also be considered a type of LR.
Under this definition, the method in~\cite{Rudin:13} is formally equivalent to the existing method~\cite{Kikuchi:19} described in Sect.~\ref{sec:introduction}.
Although the approaches described above all mitigate the low-frequency problem, none address the zero-frequency problem.

Many probability estimation approaches have attempted to address low- and zero-frequency problems through the use of smoothing techniques, which discount probability estimates for observed elements and distribute them for unobserved elements.
Here, we focus on smoothing for a probability $p(w)$ of an $N$-gram $w$~\cite{Jelinek:80,Katz:87,Witten:91,Ney:91,Ney:94,MacKay:95,Kneser:95}.
This type of smoothing uses lower order $N$-gram (partial elements of $w$) frequencies in addition to original $N$-gram frequencies.
Even $p(w)$ for unobserved $w$ can be estimated by using them alternatively.
Although $w$-decomposition is related to our approach, these techniques are not directly applicable to LR estimation.
Because the LR is given as a ratio of probability distributions, representing it as a ratio of smoothed estimates appears to be a natural approach;
however, we would expect that this approach would still face the low-frequency problem as a result of indirect estimation and, therefore, we must verify it experimentally.

Other approaches for mitigating the two problems have been used to estimate odds ratios, which are represented as ratios of probability distributions, for correcting individual probability estimators.
In~\cite{Parzen:02} and~\cite{Raweesawat:16}, unbiased and Bayes estimators were used for cell probability estimation.
These estimators reduce the estimation bias for rare elements and can avoid the zero-frequency problem in cases in which all observed elements are infrequent.
However, we treat elements ranging from high- to zero-frequency in LR estimation.
Moreover, approaches involving ratios of corrected estimates have been found to face the low-frequency problem when used for LR estimation~\cite{Kikuchi:19}, which leads us to suspect that the smoothing approach described above also faces the low-frequency problem.

Some classifiers decompose item sequences.
For instance, naive Bayes classifiers are based on a strong (naive) independence assumption and Bayes' theorem.
The classifiers approximate the probability $p(D|C)$ that a document $D=a_1a_2 \cdots a_N$ belongs to a class $C$ as follows:
\begin{align*}
p(D|C) \approx \prod_i^N p(a_i|C),
\end{align*}
where $a_i$ is an item, e.g., a word or letter, in $D$ and $p(a_i|C)$ is estimated using smoothing techniques.
In this approach, the low- and zero-frequency problems are mitigated.
The approximation is based on a between-item conditional independence assumption; i.e., $p(a_i|C, a_j)=p(a_i|C)$, where $i \neq j$.
This assumption, however, often does not hold in real-world data,
and extended methods to alleviate it have been proposed~\cite{Frank:02,Jiang:12,Wu:16,Tang:16,Diab:17,Jiang:18}.
Our approach for itemizing $N$-grams and incorporating dependency between items was inspired by this framework.

\section{Direct Estimation of Likelihood Ratio: Existing Methods}\label{sec:Direct_Estimation}

Several direct LR estimation methods, including the use of kernel mean matching~\cite{Huang:07}, logistic regression~\cite{Bickel:07}, Kullback-Leibler divergence minimization~\cite{Sugiyama:08}, and least-squares~\cite{Kanamori:09} approaches, have been proposed.
These methods are used to estimate LRs represented  by continuous distributions.
A method for estimating them represented by discrete distributions, was proposed by~\cite{Kikuchi:19}.
This method is a variant of the least-squares approach called {\it unconstrained Least-Squares Importance Fitting} (uLSIF).
Our proposed method, described in Sect.~\ref{sec:Proposed_Mothod}, follows the uLSIF framework, although we use the variant to treat discrete elements.
Thus, in this Sect. we explain the problem definition and then describe the original uLSIF and its variant.

\subsection{Problem Definition}\label{sec:Problem_Definition}

We define the overall data domain as $D$.
In Sects.~\ref{sec:uLSIF} and \ref{sec:Kikuchi19}, the domains are defined as $D \subset \mathbb{R}^d$ and $D \subset \mathbb{U}$, respectively.
$\mathbb{R}^d$ is a real $d$-dimensional space.
$\mathbb{U}$ is a set with $v$ discrete elements and is also called a finite alphabet in information theory.
We examine a case in which we obtain an independent and identically distributed (i.i.d.) sample $\{x_{i}^{\rm de}\}_{i=1}^{n_{\rm de}}$ from a probability distribution with density $p_{\rm de}(x)$ and an i.i.d. sample $\{x_{j}^{\rm nu}\}_{j=1}^{n_{\rm nu}}$ from a probability distribution with density $p_{\rm nu}(x)$:
\begin{align*}
\{x_{i}^{\rm de}\}_{i=1}^{n_{\rm de}} \overset{\rm i.i.d.}{\sim} p_{\rm de}(x), \quad
\{x_{j}^{\rm nu}\}_{j=1}^{n_{\rm nu}} \overset{\rm i.i.d.}{\sim} p_{\rm nu}(x).
\end{align*}
Following previous studies, we assume that $p_{\rm de}(x)$ satisfies the condition
\begin{align*}
p_{\rm de}(x) > 0 \quad {\rm for\ all\ } x \in D,
\end{align*}
which makes it possible to define an LR for all $x$.
We then attempt to estimate the following LR $r(x)$ directly from two samples, $\{x_{i}^{\rm de}\}_{i=1}^{n_{\rm de}}$ and $\{x_{j}^{\rm nu }\}_{j=1}^{n_{\rm nu}}$:
\begin{align*}
r(x) = \frac{p_{\rm nu}(x)}{p_{\rm de}(x)}.
\end{align*}

\subsection{Unconstrained Least-Squares Importance Fitting (uLSIF)}\label{sec:uLSIF}

In uLSIF~\cite{Kanamori:09}, $r(x)$ is modeled by the following linear model:
\begin{align}
\label{eq:Linear_model}
\widehat{r}(x) = \sum_{l=1}^{b} \beta_l \varphi_l (x),
\end{align}
where $\mbox{\boldmath $\beta$}=(\beta_1, \beta_2, \ldots ,\beta_{b})^{\mathrm{T}}$ are the parameters learned from data samples and $\{\varphi_l\}_{l=1}^{b}$ are the basis functions, which are always non-negative.
Note that $b$ and $\{\varphi_l\}_{l=1}^{b}$ are independent of the samples $\{x_{i}^{\rm de}\}_{i=1}^{n_{\rm de}}$ and $\{x_{i}^{\rm nu }\}_{i=1}^{n_{\rm nu}}$.
The goal of uLSIF is to solve the optimization problem\footnote{The objective function in Eq.(\ref{eq:uLSIF}) is derived from the square loss of $\widehat{r}(x)$ and $r(x)$. See Appendix A for a derivation of the first two terms in this function.}:
\begin{align}
\label{eq:uLSIF}
\min_{\mbox{\boldmath $\beta$} \in \mathbb{R}^b} \left[\frac{1}{2} \mbox{\boldmath $\beta$}^{\mathrm{T}} \widehat{\mbox{\boldmath $H$}} \mbox{\boldmath $\beta$} - \widehat{\mbox{\boldmath $h$}}^{\mathrm{T}} \mbox{\boldmath $\beta$} + \frac{\lambda}{2} \mbox{\boldmath $\beta$}^{\mathrm{T}} \mbox{\boldmath $\beta$}\right],
\end{align}
where $\widehat{\mbox{\boldmath $H$}}$ is a $b \times b$ matrix in which the $(l,l')$-th element is given by
\begin{align}
\label{eq:H_hat}
\widehat{H}_{l,l'}=\frac{1}{n_{\rm de}}\sum_{i=1}^{n_{\rm de}} \varphi_l(x_i^{\rm de}) \varphi_{l'}(x_i^{\rm de}),
\end{align}
and $\widehat{\mbox{\boldmath $h$}}$ is a $b$-dimensional vector in which the $l$-th element is given by
\begin{align}
\label{eq:h_hat}
\widehat{h}_l=\frac{1}{n_{\rm nu}}\sum_{j=1}^{n_{\rm nu}} \varphi_l(x_j^{\rm nu}).
\end{align}
In Eq.(\ref{eq:uLSIF}), a penalty term $\frac{\lambda}{2} \mbox{\boldmath $\beta$}^{\mathrm{T}} \mbox{\boldmath $\beta$}$ is introduced for regularization to $\mbox{\boldmath $\beta$}$;
$\lambda \ (\geq 0)$ is a regularization parameter and $\mbox{\boldmath $\beta$}^{\mathrm{T}} \mbox{\boldmath $\beta$}/2$ is an $\ell_2$-regularizer.
As the objective function in Eq.(\ref{eq:uLSIF}) is a convex quadratic function, its solutions can be written as
\begin{align*}
\widetilde{\mbox{\boldmath $\beta$}}(\lambda)=(\widehat{\mbox{\boldmath $H$}}+\lambda\mbox{\boldmath $1$}_b)^{-1}\widehat{\mbox{\boldmath $h$}},
\end{align*}
where $\mbox{\boldmath $1$}_b$ is a $b$-dimensional vector containing all ones.
As the optimization problem does not have non-negativity constraints on the parameters, some of them can be negative.
Therefore, $\widetilde{\mbox{\boldmath $\beta$}}(\lambda)$ are modified as follows by taking into account the non-negativity of $r(x)$:
\begin{align*}
\widehat{\mbox{\boldmath $\beta$}}(\lambda)={\rm max}(\mbox{\boldmath $0$}_b, \widetilde{\mbox{\boldmath $\beta$}}(\lambda)),
\end{align*}
where the ``max'' operation is applied in the element-wise manner and $\mbox{\boldmath $0$}_b$ is a $b$-dimensional vector containing all zeros.
$\widehat{\mbox{\boldmath $\beta$}}(\lambda)$ are the final solutions of Eq.(\ref{eq:uLSIF}). 

uLSIF uses the structure of a sample space to model $r(x)$.
For this reason, the selection of basis functions is important.
Moreover, the $\ell_2$-regularizer gives prior information that $\mbox{\boldmath $\beta$}$ are uniform under the condition that no samples are observed from the space.

\subsection{A Variant of uLSIF for Discrete Distributions}
\label{sec:Kikuchi19}

The primary alteration to the original uLSIF involves a change of basis functions.
In natural language processing (NLP), observed samples often comprise discrete elements such as words and letters rather than continuous elements.
uLSIF uses Gaussian kernels in its basis functions,
but these cannot take into account the discreteness of the sample space.
To apply uLSIF in discrete cases, Kikuchi et al.~\cite{Kikuchi:19} proposed the use of the basis functions $\{\varphi_l\}_{l=1}^{v}$:
\begin{align}
\label{eq:phi}
\varphi_l(x)= 
\begin{cases}
    1 & x = x_{(l)}, \\
    0 & {\rm otherwise}.
  \end{cases}
\end{align}
In the above, $x$ is a discrete element,
$v$ is the number of possible element types, and $l$ is an index specifying the specific type from among $v$ types; that is, $x_{(l)}$ indicates the $l$-th element among $v$ different elements.
The functions can also be regarded as Kronecker delta functions.
In addition, $r(x_{(m)})$, $1 \leq m \leq v$, is modeled as Eq.(\ref{eq:Linear_model}):
\begin{align*}
\widehat{r}_{\rm K}(x_{(m)}) &= \sum_{l=1}^{v} \beta_l \varphi_l (x_{(m)}) = \beta_m.
\end{align*}
As the basis function $\varphi_l(x)$ becomes zero unless $x$ is identical to $x_{(l)}$, 
the $\widehat{H}_{l, l'}$ and $\widehat{h}_l$ corresponding to Eqs.(\ref{eq:H_hat}) and (\ref{eq:h_hat}), respectively, are given by
\begin{align}
\widehat{H}_{l,l'} &= \begin{cases}
	\frac{1}{n_{\rm de}} \sum_{i=1}^{n_{\rm de}} \varphi_l (x_i^{\rm de}) \varphi_{l'} (x_i^{\rm de}) & (l=l') \\
	0 & (l \neq l')
	\end{cases} \nonumber \\
&=\frac{1}{n_{\rm de}} c_{\rm de} (x_{(l)}) \quad (l=l'), \nonumber \\
\widehat{h}_l &= \frac{1}{n_{\rm nu}} \sum_{j=1}^{n_{\rm nu}} \varphi_l (x_j^{\rm nu}) = \frac{1}{n_{\rm nu}} c_{\rm nu} (x_{(l)}), \nonumber
\end{align}
where $c_*(x_{(l)})$, $* \in \{\rm de, nu\}$ is the frequency of $x_{(l)}$ sampled from a probability distribution with density $p_*(x_{(l)})$ and
$\widehat{\mbox{\boldmath $H$}}$ is a $v \times v$ diagonal matrix.
From the above, $\widehat{r}_{\rm K}(x_{(m)})$ can be calculated as
\begin{align}
\label{eq:solution_Kikuchi19}
\widehat{r}_{\rm K}(x_{(m)}) &= \widetilde{\beta}_m (\lambda) \nonumber \\
&= (\widehat{H}_{m,m} + \lambda)^{-1} \widehat{h}_m \nonumber \\
&= \left(\frac{1}{n_{\rm de}} c_{\rm de} (x_{(m)}) + \lambda \right)^{-1} \frac{1}{n_{\rm nu}} c_{\rm nu} (x_{(m)}),
\end{align}
which is a non-negative value and therefore becomes the solution\footnote{As  the original paper~\cite{Kikuchi:19} was written in Japanese, in Sect.~\ref{sec:Regularization_K} we present an estimation example for further clarification.}.
The regularization parameter $\lambda$ results in an underestimation of $r(x_{(m)})$.
If $\lambda$ is zero, $\widehat{r}_{\rm K}(x_{(m)})$ is $r_{\rm MLE}(w)=\frac{\widehat{p}_{\rm nu}(x_{(m)})}{\widehat{p}_{\rm de}(x_{(m)})}$, which is equivalent to Eq.(\ref{eq:Indirect_MLE}) with $\widehat{p}_*(x_{(m)})$ representing the MLE\footnotemark[1] of a probability distribution with density $p_*(x_{(m)})$.
Eq.(\ref{eq:solution_Kikuchi19}) adjusts only the denominator $\frac{1}{n_{\rm de}} c_{\rm de} (x_{(m)})$ of $r_{\rm MLE}(w)$ using $\lambda$.

\subsection{Regularization Effect on $\widehat{r}_{\rm K}(x_{(m)})$}\label{sec:Regularization_K}

\begin{table}[!t]
\caption{Frequency examples (identical to Table~\ref{tab:Frequencies_example}).}
\renewcommand{\arraystretch}{1.3}
\label{tab:Frequencies_K}
\centering
	\begin{tabular}{ c  r  r  r  r  r } \hline
		\multicolumn{1}{ c }{$N$-gram} & \multicolumn{4}{ c }{Frequency}  & \multirow{2}{*}{$r_{\rm MLE}(w)$} \\ \cline{2-5}
		$w$ & \multicolumn{1}{ c }{$n_{\rm de}$} & \multicolumn{1}{ c }{$c_{\rm de}(w)$} & \multicolumn{1}{ c }{$n_{\rm nu}$} & \multicolumn{1}{ c }{$c_{\rm nu}(w)$} & \\ \hline \hline
		$w_{\rm A}$ & $10^7$ & 5,000 & $10^4$ & 100 & 20 \\
		$w_{\rm B}$ & $10^7$ & 50 & $10^4$ & 1 & 20 \\
		$w_{\rm C}$ & $10^7$ & 50 & $10^4$ & 2 & 40 \\
		$w_{\rm D}$ & $10^7$ & 14 & $10^4$ & 0 & 0 \\ \hline
	\end{tabular}
\end{table}
\begin{figure}[!t]
  \centering
  \includegraphics[keepaspectratio, scale=0.42]{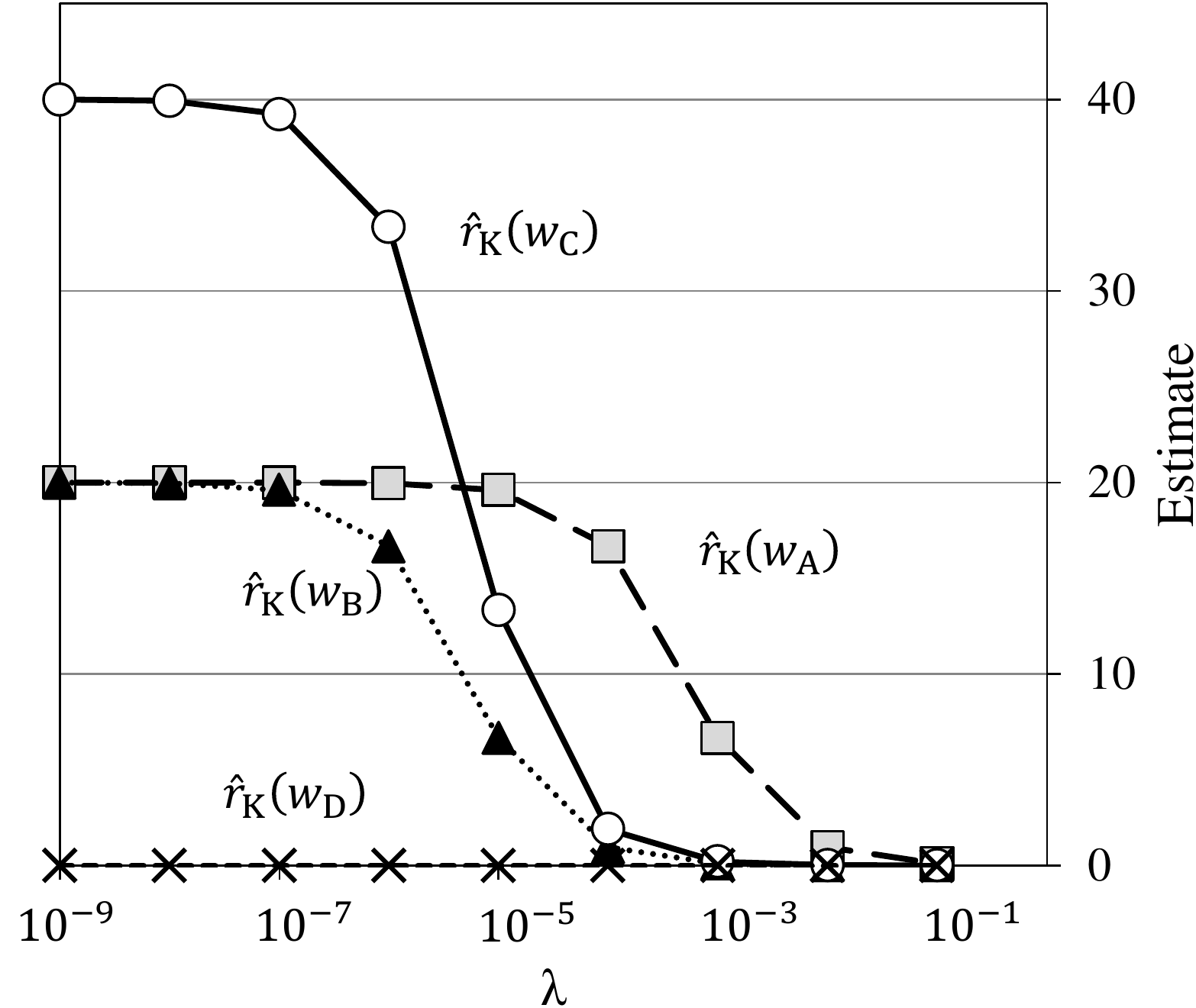}
  \caption{
  Regularization effect on $\widehat{r}_{\rm K}(w)$ given in Eq.(\ref{eq:solution_Kikuchi19}).
  The horizontal and vertical axes of the graph represent the regularization parameter $\lambda$ and estimate of $\widehat{r}_{\rm K}(w)$, respectively.
  }
  \label{fig:K}
\end{figure}

As we use the estimator $\widehat{r}_{\rm K}(x_{(m)})$ in our proposed method, understanding its function will be important.
Accordingly, in this Sect. we explain $\widehat{r}_{\rm K}(x_{(m)})$ from the perspective of regularization.
To do so, we assume that each discrete element $x$ is an $N$-gram $w$.
As in Sect.~\ref{sec:introduction}, we calculate the likelihood ratio, $r(w)=\frac{p_{\rm nu}(w)}{p_{\rm de}(w)}$, to predict the left context of NEs.
$p_{\rm de}(w)$ is the probability density where $w$ occurs at any position in the corpus, and $p_{\rm nu}(w)$ is the probability density where $w$ occurs on the left side of the NEs.
That is, if $w$ is the left context of the NEs, $r(w)$ is high. Otherwise, $r(w)$ is low.
We also assume that we observe the frequencies shown in Table~\ref{tab:Frequencies_K} (identical to Table~\ref{tab:Frequencies_example}) and use them to estimate $r(w)$ for each $w$.
Here, $c_{\rm de}(w)$ is the frequency of $w$ in the entire corpus, and $c_{\rm nu}(w)$ is the frequency of $w$ on the left side of NEs.
$\widehat{r}_{\rm K}(w)$ with $\lambda=0$ is equal to $r_{\rm MLE}(w)$ as defined by Eq.(\ref{eq:Indirect_MLE}).
Fig.~\ref{fig:K} shows the estimated $\widehat{r}_{\rm K}(w)$ as the regularization parameter $\lambda$ is varied over the range $10^{-9}, 10^{-8}, \ldots , 10^{-1}$.

At $\lambda=10^{-9}$, each $\widehat{r}_{\rm K}(w)$ is nearly equal to $r_{\rm MLE}(w)$ in Table~\ref{tab:Frequencies_K} as a result of the weak regularization.
At nearly $\lambda=10^{-7}$, $\widehat{r}_{\rm K}(w_{\rm B})$ and $\widehat{r}_{\rm K}(w_{\rm C})$ both begin to decrease,
while $\widehat{r}_{\rm K}(w_{\rm A})$ remains close to 20.
The frequencies of $w_{\rm B}$ and $w_{\rm C}$ are lower than those of $w_{\rm A}$,
indicating that the regularization effect acts more strongly on $\widehat{r}_{\rm K}(w_{\rm B})$ and $\widehat{r}_{\rm K}(w_{\rm C})$ even when $r_{\rm MLE}(w_{\rm A}) = r_{\rm MLE}(w_{\rm B})$.
As $\lambda$ increases further, $\widehat{r}_{\rm K}(w_{\rm A})$ begins to decrease, and all estimators go to zero,
suggesting that, for all $w$, $\widehat{r}_{\rm K}(w)$ becomes robust for frequencies with large $\lambda$.
Therefore, if we use $\widehat{r}_{\rm K}(w)$ in the prediction task of the left $N$-grams of NEs, we can preferentially predict high-frequency left $N$-grams, and after that we can predict low-frequency left $N$-grams.
Note that $\widehat{r}_{\rm K}(w_{\rm D})$ is always zero.
This method addresses the low-frequency problem but not the zero-frequency problem.
When dealing with high-order (long) $N$-grams, there are many $w$s having $c_{\rm de}(w)=0$ or $c_{\rm nu}(w)=0$.
Therefore, in the next section, we propose a new estimator that can address the zero-frequency problem by also using the composed items of $w$.

\section{Proposed Likelihood Ratio Estimator}\label{sec:Proposed_Mothod}

Our proposed method follows the uLSIF framework.
Our framework models LR as a linear sum of basis functions and their weighting parameters, which are optimized using least-squares minimization.
Note that we select the basis functions applicable to discrete elements, as well as the method described in Sect.~\ref{sec:Kikuchi19}.
We briefly explain our estimation procedure as follows.
For an ``itemizable'' element $w$, $r(w)$ is first roughly approximated as $r_{\rm item}(w)$ using the component frequencies of $w$ under the assumption of independence between components.
Then, the term $t_{\rm d}(w)$ based on the frequency of $w$ is added to $\widehat{r}_{\rm item}(w)$ for independence alleviation.
Finally, an estimate of $t_{\rm d}(w)$ for which the square loss of $\widehat{r}_{\rm item}(w)+\widehat{t}_{\rm d}(w)$ and $r(w)$ is minimized is made.
Below, we describe our method in detail.

\subsection{Overall Framework}\label{sec:Overall_Framework}

We examine a case in which we observe $N$-grams as itemizable elements.
Each $N$-gram $w$ is a continuous sequence of $N$ items $\{a_{k}\}_{k=1}^{N}$:
\begin{align*}
w=a_1a_2 \cdots a_N,
\end{align*}
where $a_k$ is a discrete item at the $k$-th position in $w$.
The items can be words, letters, etc.
In particular, for $N=1,2,3$, $w$ is called a unigram, bigram, and trigram, respectively.
Before going further, we define the two operators $\Psi_k$ and $\Psi_{k(l)}$ for simplicity.
These generate item patterns from $w$.

\begin{defi}[Operator $\Psi_k$]
An operator $\Psi_k$ replaces all items except $a_k$ with a wildcard symbol $\bullet$.
This symbol allows any item type match at the $k$-th position.
If $w$ is $a_1a_2a_3$, $\Psi_kw$ $(k=1,2,3)$ is $\Psi_1w=a_1 \bullet \bullet$, $\Psi_2w=\bullet a_2 \bullet$, and $\Psi_3w=\bullet \bullet a_3$.
\end{defi}

\begin{defi}[Operator $\Psi_{k(l)}$]
An operator $\Psi_{k(l)}$ replaces all items except $a_k$ with $\bullet$ and $a_k$ with $a_{(l)}$, $1 \leq l \leq v$, where $l$ is an index specifying an item type and $v$ is the number of item types that can exist; that is, $a_{(l)}$ indicates the $l$-th item among all different $v$ items.
If $w$ is $a_1a_2a_3$, $\Psi_{k(l)}w$ $(k=1,2,3)$ is $\Psi_{1(l)}w=a_{(l)} \bullet \bullet$, $\Psi_{2(l)}w=\bullet a_{(l)} \bullet$, and $\Psi_{3(l)}w=\bullet \bullet a_{(l)}$.
\end{defi}

Our initial approach is to assume the statistical independence of items $\{a_{k}\}_{k=1}^{N}$ in $w$ and to itemize $w$ into item units.
Based on this, $r(w)$ can be represented by $r_{\rm item}(w)$, which is the product of the LRs of units $\{\Psi_kw\}_{k=1}^N$:
\begin{align}
\label{eq:itemized}
r_{\rm item}(w) =& \prod_{k=1}^{N} r(\Psi_kw) \\
=& r(a_1 \bullet \bullet \cdots \bullet) r(\bullet a_2 \bullet \cdots \bullet) \nonumber \\
& \cdots r(\bullet \bullet \cdots \bullet a_N). \nonumber
\end{align}
All of $\Psi_kw$ match many $N$-grams including $w$.
Therefore, if $r(w)$ can be defined, $r(\Psi_kw)$ can also be defined.
By itemizing each $w$---even those sequences not observed in samples---we can estimate $r_{\rm item}(w)$ based on individual unit frequencies.
However, the independence assumption is inaccurate and might cause a significant estimation error.
Therefore, we add a dependency term $t_{\rm d}(w)$ to Eq.(\ref{eq:itemized}) to incorporate the dependency between items.
Our overall model is then:
\begin{align}
\label{eq:our_model}
r_{\rm ours}(w) &= r_{\rm item}(w) + t_{\rm d}(w),
\end{align}
which is estimated in two steps.
We first estimate $r_{\rm item}(w)$ using each unit frequency $c_*(\Psi_kw)$, $* \in \{\rm de ,nu\}$, and we then estimate $t_{\rm d}(w)$ using each frequency $c_*(w)$.
$t_{\rm d}(w)$ is defined for each $N$-gram $w$ and estimated so that the squared loss between our model $r_{\rm ours}(w)$ and the ideal $r(w)$ is minimized.
When at least $c_{\rm de}(w)>0$ or $c_{\rm nu}(w)>0$ and the estimate of $r_{\rm item}(w) + t_{\rm d}(w)$ is non-negative, we can obtain the exact solution that minimizes the squared error.

\subsection{Estimation of Itemized LR $r_{\rm item}(w)$}\label{sec:Itemized_LR}

Given an $N$-gram $w$, we estimate $r(\Psi_kw)$ in Eq.(\ref{eq:itemized}) by applying the method described in Sect.~\ref{sec:Kikuchi19}.
We define each of the basis functions $\{\varphi_l(\Psi_kw)\}_{l=1}^{v}$ by
\begin{align*}
\varphi_l(\Psi_kw)=
\begin{cases}
    1 & \Psi_kw = \Psi_{k(l)}w, \\
    0 & {\rm otherwise},
\end{cases}
\end{align*}
where $l$ is the index specifying an item type that is not replaced by the wildcard symbol $\bullet$ and
$v$ is the number of item types that can exist.
If $\Psi_kw$ is identical to $\Psi_{k(m)}w$, $1 \leq m \leq v$, that is, if $a_k$ is identical to $a_{(m)}$, $\widehat{r}_{\rm K}(\Psi_{k(m)}w)$ can be calculated as
\begin{align*}
\widehat{r}_{\rm K}(\Psi_{k(m)}w) =& \left(\frac{1}{n_{\rm de}} c_{\rm de} (\Psi_{k(m)}w) + \lambda_{\rm item} \right)^{-1} \\
& \times \frac{1}{n_{\rm nu}} c_{\rm nu} (\Psi_{k(m)}w),
\end{align*}
where $\lambda_{\rm item} \ (\geq 0)$ is a regularization parameter and
$n_{\rm de}$ and $n_{\rm nu}$ are the total frequencies of all $N$-grams observed from samples $\{w_{i}^{\rm de}\}_{i=1}^{n_{\rm de}}$ and $\{w_{j}^{\rm nu }\}_{j=1}^{n_{\rm nu}}$, respectively.
From the above, we estimate $r_{\rm item}(w)$ as the product of $\{\widehat{r}_{\rm K}(\Psi_kw)\}_{k=1}^N$:
\begin{align}
\label{eq:itemized_estimated}
\widehat{r}_{\rm item}(w) &= \prod_{k=1}^{N} \widehat{r}_{\rm K}(\Psi_kw).
\end{align}
In our experiments, $n_{\rm de}$ and $n_{\rm nu}$ are the total frequencies of $N$-grams in the training set and on the left side of the NEs, respectively.
From these definitions, $n_{\rm de}>>n_{\rm nu}$ holds.
Additionally, many $N$-grams comprise low-frequency items.
Considering these, $\widehat{p}_{\rm de}(\Psi_kw)<<\widehat{p}_{\rm nu}(\Psi_kw)$ holds for most $\Psi_kw$.
The value of $\widehat{p}_{\rm de}(\Psi_kw)$ is strongly reflected in the estimate of $r(\Psi_kw)$. Therefore, we correct the denominator of $r(\Psi_kw)$.
$\widehat{r}_{\rm item}(w)$ has $N$ regularization parameters.
However, searching for the optimal combination of all $\lambda_{\rm item}$ is computationally expensive.
On the other hand, $\widehat{p}_{\rm de}(\Psi_kw)$ is the occurrence probability of $\Psi_kw $ in the training set.
Here, if the words $a_k$ and $a_{k'}$ are identical, then $\widehat{p}_{\rm de}(\Psi_kw) \approx \widehat{p}_{\rm de}(\Psi_{k'}w)$, $k \ne {k'}$.
That is, $\widehat{p}_{\rm de}(\Psi_kw)$ is almost determined by the word type of $a_k$, not $k$.
For these reasons, we consider them as common parameters and set the parameters all to the same value.

\subsection{Estimation of Dependency Term $t_{\rm d}(w)$.}

Given an $N$-gram $w$ and $\widehat{r}_{\rm item}(w)$, we estimate $t_{\rm d}(w)$ in Eq.(\ref{eq:our_model}).
We model $t_{\rm d}(w)$ as the following linear model:
\begin{align*}
\widehat{t}_{\rm d}(w) = \sum_{l=1}^v \beta_l \varphi_l(w),
\end{align*}
where $\mbox{\boldmath $\beta$}=(\beta_1, \beta_2, \ldots ,\beta_{v})^{\mathrm{T}}$ are the parameters learned from data samples and $\{\varphi_l(w)\}_{l=1}^{v}$ are the basis functions, which are always non-negative.
Similar to Eq.(\ref{eq:phi}), we define each of the basis functions for $w$ as
\begin{align*}
\varphi_l(w)= 
\begin{cases}
    1 & w = w_{(l)}, \\
    0 & {\rm otherwise},
\end{cases}
\end{align*}
where $l$ is the index specifying a particular $N$-gram type from among all $N$-gram types; that is, $w_{(l)}$ indicates the $l$-th $N$-gram from among all $N$-grams.
Note that $v$ is not defined as in Sect.~\ref{sec:Itemized_LR} but here represents the number of possible $N$-gram types.
For identical $w$ with $w_{(m)}$, $1 \leq m \leq v$, our model $r_{\rm ours}(w_{(m)})$ is given as follows:
\begin{align}
\label{eq:our_model_estimated}
r_{\rm ours}(w_{(m)}) &= \widehat{r}_{\rm item}(w_{(m)}) + \widehat{t}_{\rm d}(w_{(m)}) \nonumber \\
&= \widehat{r}_{\rm item}(w_{(m)}) + \beta_{m}.
\end{align}
$r_{\rm item}(w_{(m)})$ is replaced by $\widehat{r}_{\rm item}(w_{(m)})$, as shown in Eq.(\ref{eq:itemized_estimated}), and treated as a constant.
To determine $\mbox{\boldmath $\beta$}$ from the data samples, we solve the following problem:
\begin{align}
\label{eq:our_problem}
\min_{\mbox{\boldmath $\beta$} \in \mathbb{R}^v} \left[ \widehat{J}(\mbox{\boldmath $\beta$}) + \frac{\lambda_{\rm d}}{2} \mbox{\boldmath $\beta$}^{\mathrm{T}} \mbox{\boldmath $\beta$} \right],
\end{align}
where $\lambda_{\rm d} \ (\geq 0)$ is a regularization parameter and $\widehat{J}(\mbox{\boldmath $\beta$})$ are the squared-loss-based terms\footnote{See Appendix B for the derivation of Eq.(\ref{eq:our_J_hat}).}:
\begin{align}
\label{eq:our_J_hat}
\widehat{J}(\mbox{\boldmath $\beta$}) =& \frac{1}{2}\sum_{l=1}^v {\beta_l}^2 \left( \frac{1}{n_{\rm de}} \sum_{i=1}^{n_{\rm de}} \varphi_l(w_i^{\rm de}) \right) \nonumber \\
&+ \widehat{r}_{\rm item}(w_{(m)}) \sum_{l=1}^v \beta_l \left( \frac{1}{n_{\rm de}} \sum_{i=1}^{n_{\rm de}} \varphi_l(w_i^{\rm de}) \right) \nonumber \\
&- \sum_{l=1}^v \beta_l \left( \frac{1}{n_{\rm nu}} \sum_{j=1}^{n_{\rm nu}} \varphi_l(w_j^{\rm nu}) \right).
\end{align}
As the objective function in Eq.(\ref{eq:our_problem}) is a convex quadratic function, we can differentiate it with respect to $\beta_m$ and set it to zero:
\begin{align*}
& \frac{\partial}{\partial \beta_m} \left( \widehat{J}(\mbox{\boldmath $\beta$}) + \frac{\lambda_{\rm d}}{2} {\mbox{\boldmath $\beta$}}^{\mathrm{T}} \mbox{\boldmath $\beta$} \right) \\
=& \beta_m \left( \frac{1}{n_{\rm de}} \sum_{i=1}^{n_{\rm de}} \varphi_m(w_i^{\rm de}) + \lambda_{\rm d} \right) + \\
&\widehat{r}_{\rm item}(w_{(m)}) \left( \frac{1}{n_{\rm de}} \sum_{i=1}^{n_{\rm de}} \varphi_m(w_i^{\rm de}) \right) - \frac{1}{n_{\rm nu}} \sum_{j=1}^{n_{\rm nu}} \varphi_m(w_j^{\rm nu}) \\
=& 0.
\end{align*}
Solving the above for $\beta_m$, we obtain
\begin{align}
\label{eq:tilda_beta}
& \widetilde{\beta}_m(\lambda_{\rm d}) \nonumber \\
=& \left( \frac{1}{n_{\rm de}} \sum_{i=1}^{n_{\rm de}} \varphi_m(w_i^{\rm de}) + \lambda_{\rm d} \right)^{-1} \times \nonumber \\
& \left\{ \frac{1}{n_{\rm nu}} \sum_{j=1}^{n_{\rm nu}} \varphi_m(w_j^{\rm nu}) - \widehat{r}_{\rm item}(w_{(m)}) \left( \frac{1}{n_{\rm de}} \sum_{i=1}^{n_{\rm de}} \varphi_m(w_i^{\rm de}) \right) \right\} \nonumber \\
=& \left( \frac{c_{\rm de}(w_{(m)})}{n_{\rm de}}  + \lambda_{\rm d} \right)^{-1} \times \nonumber \\
& \left\{ \frac{c_{\rm nu}(w_{(m)})}{n_{\rm nu}} - \widehat{r}_{\rm item}(w_{(m)}) \left( \frac{c_{\rm de}(w_{(m)})}{n_{\rm de}} \right) \right\}.
\end{align}
This is the solution of the optimization problem shown in Eq.(\ref{eq:our_problem})\footnote{As shown in Eq.(\ref{eq:tilda_beta}), $\widehat{t}_{\rm d}(w_{(m)})$ is calculated from word-based frequencies, and we set the largest $N$-gram order $N$ to 4 in our experiments. Therefore, it is feasible to estimate $t_{\rm d}$ within a realistic time.}.
Substituting $\widetilde{\beta}_m(\lambda_{\rm d})$ into Eq.(\ref{eq:our_model_estimated}), our model can be calculated as
\begin{align*}
\widetilde{r}_{\rm ours}(w_{(m)}) &= \widehat{r}_{\rm item}(w_{(m)}) + \widetilde{\beta}_m(\lambda_{\rm d}).
\end{align*}
As this can be negative because $\widetilde{\beta}_m(\lambda_{\rm d})$ includes a subtraction term, we modify $\widetilde{r}_{\rm ours}(w_{(m)})$ as
\begin{align}
\label{eq:solution_ours}
\widehat{r}_{\rm ours}(w_{(m)})={\rm max}(0, \widetilde{r}_{\rm ours}(w_{(m)}))
\end{align}
to obtain our estimator.
The estimator $\widehat{r}_{\rm ours}(w_{(m)})$ has two parameters, $\lambda_{\rm item}$ and $\lambda_{\rm d}$,
which balance the impacts of $r_{\rm item}(w)$ and $t_{\rm d}(w)$ in Eq.(\ref{eq:our_model}) while mitigating the low-frequency problem.
As a result, the estimator allows us to control the dependency strength between items.

\subsection{Regularization Effect on $\widehat{r}_{\rm ours}(w)$}\label{sec:Regularization_ours}
\begin{table*}[!t]
\caption{Examples of original and itemized frequencies.}
\renewcommand{\arraystretch}{1.3}
\label{tab:Frequencies_ours}
\centering
	\begin{tabular}{ c  r  r  r  r  r  r  r  r  r  r  r } \hline
		\multicolumn{1}{ c }{Bigram} & \multicolumn{4}{ c }{Original Freq.}  & \multirow{2}{*}{$r_{\rm MLE}(w)$} & \multicolumn{4}{ c }{Itemized Freq.} & \multicolumn{1}{ l }{$\widehat{r}_{\rm item}(w)$} \\ \cline{2-5} \cline{7-10}
		$w=a_1a_2$ & \multicolumn{1}{ c }{$n_{\rm de}$} & \multicolumn{1}{ c }{$c_{\rm de}(w)$} & \multicolumn{1}{ c }{$n_{\rm nu}$} & \multicolumn{1}{ c }{$c_{\rm nu}(w)$} && \multicolumn{1}{ c }{$c_{\rm de}(a_1 \ \bullet)$} & \multicolumn{1}{ c }{$c_{\rm nu}(a_1 \ \bullet)$} & \multicolumn{1}{ c }{$c_{\rm de}(\bullet \ a_2)$} & \multicolumn{1}{ c }{$c_{\rm nu}(\bullet \ a_2)$} & \multicolumn{1}{ l }{with $\lambda_{\rm item}=0$} \\ \hline \hline
		$w_{\rm A}$ & $10^7$ & 5,000 & $10^4$ & 100 & 20 & 8,000 & 1,000 & 5,000 & 500 & 12,500 \\
		$w_{\rm B}$ & $10^7$ & 50 & $10^4$ & 1 & 20 & 240 & 30 & 150 & 15 & 12,500 \\
		$w_{\rm C}$ & $10^7$ & 50 & $10^4$ & 2 & 40 & 80 & 10 & 50 & 5 & 12,500\\
		$w_{\rm D}$ & $10^7$ & 14 & $10^4$ & 0 & 0 & 80 & 10 & 50 & 5 & 12,500 \\ \hline
	\end{tabular}
\end{table*}
\begin{figure}[!t]
  \centering
  \includegraphics[keepaspectratio, scale=0.46]{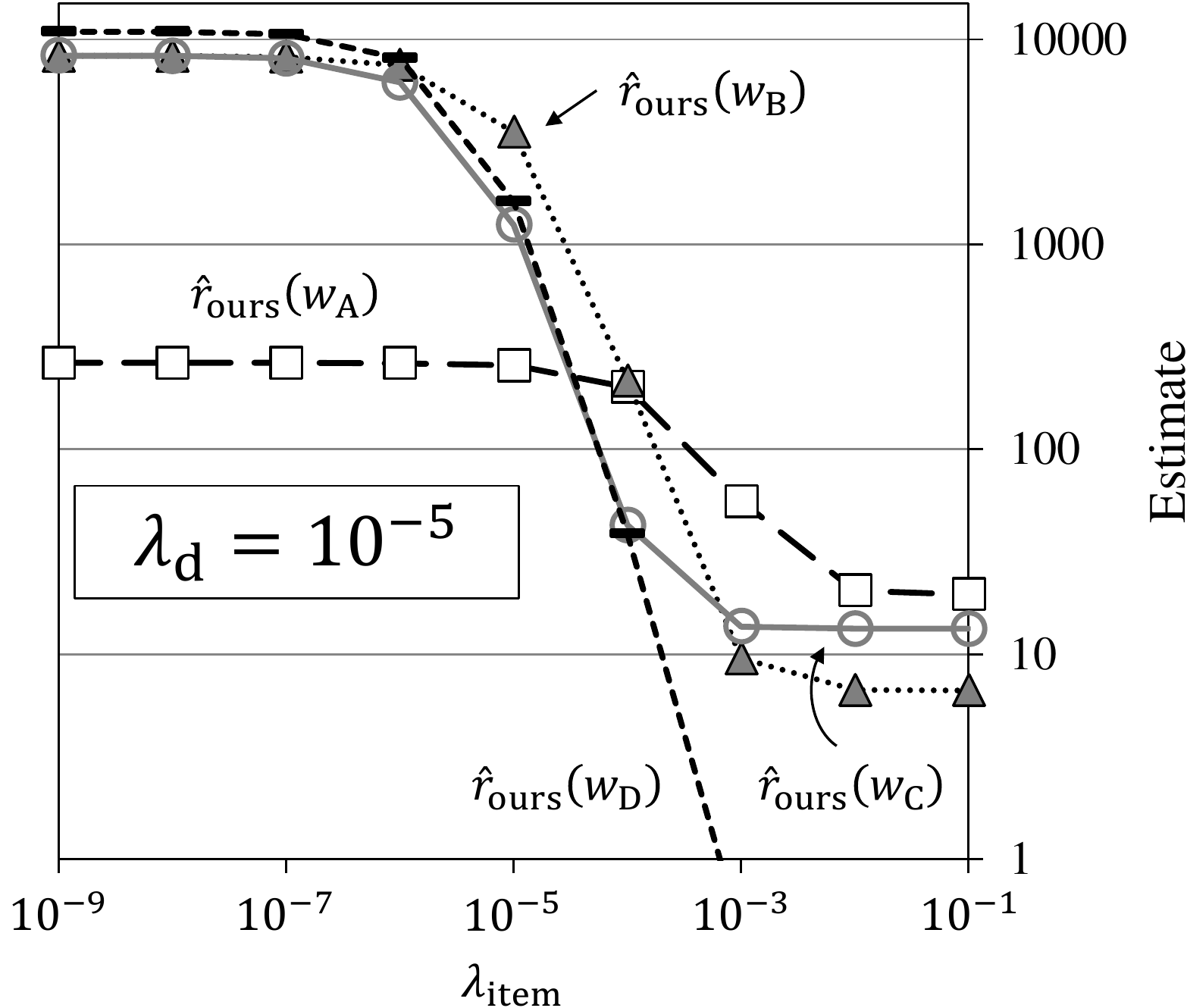}
  \caption{
  Regularization effect on our estimator $\widehat{r}_{\rm ours}(w)$ shown in Eq.(\ref{eq:solution_ours}).
  The horizontal and vertical axes of the graph represent the regularization parameter $\lambda_{\rm item}$ and estimate of $\widehat{r}_{\rm ours}(w)$, respectively.
  To confirm the effect of $\lambda_{\rm item}$, we have fixed $\lambda_{\rm d}$ at $10^{-5}$.
  }
  \label{fig:ours_item}
\end{figure}
\begin{figure}[!t]
  \centering
  \includegraphics[keepaspectratio, scale=0.46]{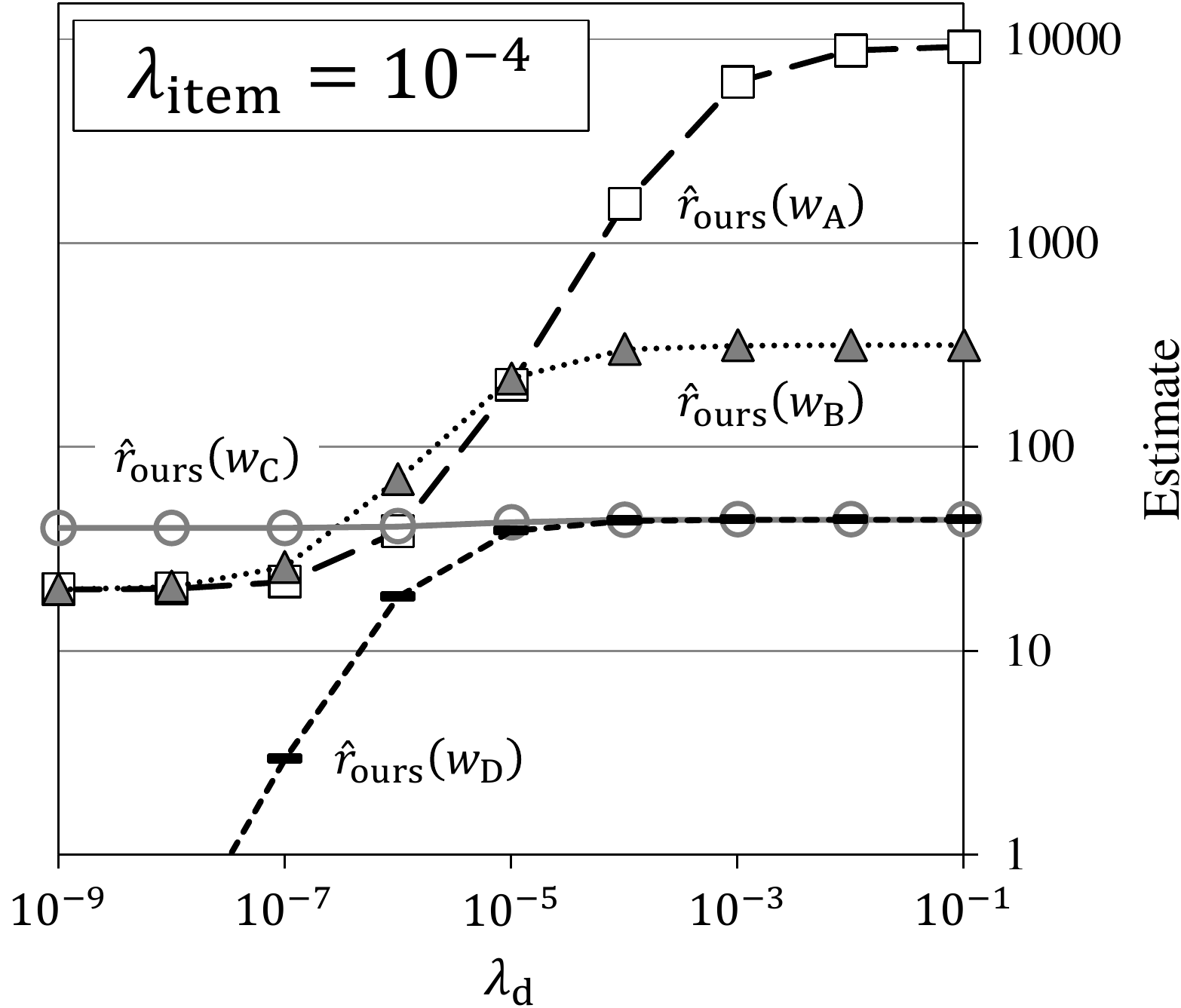}
  \caption{
  Regularization effect on our estimator $\widehat{r}_{\rm ours}(w)$ shown in Eq.(\ref{eq:solution_ours}).
  The horizontal and vertical axes of the graph represent the regularization parameter $\lambda_{\rm d}$ and estimate of $\widehat{r}_{\rm ours}(w)$, respectively.
  To confirm the effect of $\lambda_{\rm d}$, we have fixed $\lambda_{\rm item}$ at $10^{-4}$.
  }
  \label{fig:ours_d}
\end{figure}

In this section, we explain our estimator $\widehat{r}_{\rm ours}(w)$ in Eq.(\ref{eq:solution_ours}) from the perspective of the regularization process.
As in Sect.~\ref{sec:introduction}, we calculate the likelihood ratio, $r(w)=\frac{p_{\rm nu}(w)}{p_{\rm de}(w)}$, to predict the left context of the NEs.
$p_{\rm de}(w)$ is the probability density where $w$ occurs at any position in the corpus, and $p_{\rm nu}(w)$ is the probability density where $w$ occurs on the left side of the NEs.
That is, if $w$ is the left context of NEs, $r(w)$ is high. Otherwise, $r(w)$ is low.
We assume that we observe the frequencies shown in Table~\ref{tab:Frequencies_ours} and use them to estimate $r(w)$ for each $w$.
$\widehat{r}_{\rm ours}(w)$ has two regularization parameters: $\lambda_{\rm item}$ and $\lambda_{\rm d}$.
To clarify the effect of each parameter, we fix one and change the other over the range $10^{-9}, 10^{-8}, \ldots , 10^{-1}$.
Figs.~\ref{fig:ours_item} and \ref{fig:ours_d} show the regularization effects on $\widehat{r}_{\rm ours}(w)$ by $\lambda_{\rm item}$ and $\lambda_{\rm d}$, respectively.

In Fig.~\ref{fig:ours_item}, $\widehat{r}_{\rm ours}(w)$ for all $w$ decreases as $\lambda_{\rm item}$ grows.
For $\lambda_{\rm item}=10^{-9}$, $\widehat{r}_{\rm ours}(w_{\rm B})$, $\widehat{r}_{\rm ours}(w_{\rm C})$, and $\widehat{r}_{\rm ours}(w_{\rm D})$ are all nearly 10,000 because of the weak regularization.
The estimates are close to the $\widehat{r}_{\rm item}(w)$ obtained with $\lambda_{\rm item}=0$ in Table~\ref{tab:Frequencies_ours}.
We infer that $\widehat{r}_{\rm ours}(w_{\rm A})$ is lower than the other estimates as a result of the subtraction term in Eq.(\ref{eq:tilda_beta}).
For $\lambda_{\rm item}=10^{-5}$, $\widehat{r}_{\rm ours}(w_{\rm C})$ is lower than $\widehat{r}_{\rm ours}(w_{\rm B})$.
As the itemized frequencies of $w_{\rm C}$ are lower than those of $w_{\rm B}$, the regularization acts more strongly on $\widehat{r}_{\rm ours}(w_{\rm C})$.
As $\lambda_{\rm item}$ grows further, $\widehat{r}_{\rm ours}(w_{\rm A})$ produces the highest estimate, $\widehat{r}_{\rm ours}(w_{\rm B})$ and $\widehat{r}_{\rm ours}(w_{\rm C})$ produce nearly identical estimates, and $\widehat{r}_{\rm ours}(w_{\rm D})$ goes to zero.
Here $\widehat{r}_{\rm ours}(w)$ becomes robust for itemized frequencies and original frequencies $c_*(w)$ are more heavily emphasized because $\lambda_{\rm item} > \lambda_{\rm d}$.
The setting, $\lambda_{\rm item} > \lambda _{\rm d}$, is effective in the case where the required dependency between items is strong.
Let us assume that the entity type is location name, and the $N$-gram $w$ is ``Hospital in" ($a_1=``{\rm Hospital},"$ $a_2=``{\rm in}"$).
Because the preposition ``in" occurs in various contexts, and ``Hospital" at the left of ``in" is the key to predicting whether $w$ is the left context of location names.
For such an example, the co-occurrence frequency $c_{*}(w)$ of two words can be used to incorporate the dependency into the estimate.

Unlike in Fig.~\ref{fig:ours_item}, in Fig.~\ref{fig:ours_d} $\widehat{r}_{\rm ours}(w)$ increases for all $w$ as $\lambda_{\rm d}$ grows.
For $\lambda_{\rm d}=10^{-9}$, each $\widehat{r}_{\rm ours}(w)$ is close to the corresponding $\widehat{r}_{\rm MLE}(w)$ in Table~\ref{tab:Frequencies_ours} because of the weak regularization.
For $\lambda_{\rm d}=10^{-5}$, $\widehat{r}_{\rm ours}(w_{\rm B})$ is lower than $\widehat{r}_{\rm ours}(w_{\rm A})$.
As the $c_*(w_{\rm B})$ are lower than the $c_*(w_{\rm A})$, the regularization acts more strongly on $\widehat{r}_{\rm ours}(w_{\rm B})$.
As $\lambda_{\rm d}$ grows further, $\widehat{r}_{\rm ours}(w_{\rm A})$ and $\widehat{r}_{\rm ours}(w_{\rm B})$ produce the highest and second-highest estimates, respectively, while $\widehat{r}_{\rm ours}(w_{\rm D})$ produces an estimate nearly equal to that for $\widehat{r}_{\rm ours}(w_{\rm C})$ despite the fact that $c_{\rm nu}(w_{\rm D})=0$.
Here $\widehat{r}_{\rm ours}(w)$ becomes robust for original frequencies and itemized frequencies are more heavily emphasized because $\lambda_{\rm d} > \lambda_{\rm item}$.
The setting, $\lambda_{\rm d} > \lambda _{\rm item}$, is effective in the case where the required dependency between items is weak.
Let us assume that the entities type is person name, and $w=``{\rm to\ Mr.}"$
Here, we can imagine that a person name will be on the right side of $a_2=``{\rm Mr.,}"$ regardless of $a_1$.
For such an example, the frequency information for each word $a_k$ can be strongly reflected in the estimate.

Overall, these results indicate a trade-off between the original and itemized frequencies; as such, we can control the dependency between items by tuning $\lambda_{\rm item}$ and $\lambda_{\rm d}$.

\section{Objective Evaluation}\label{sec:Objective_Evaluation}

\begin{figure}[!t]
  \centering
  \includegraphics[keepaspectratio, scale=0.45]{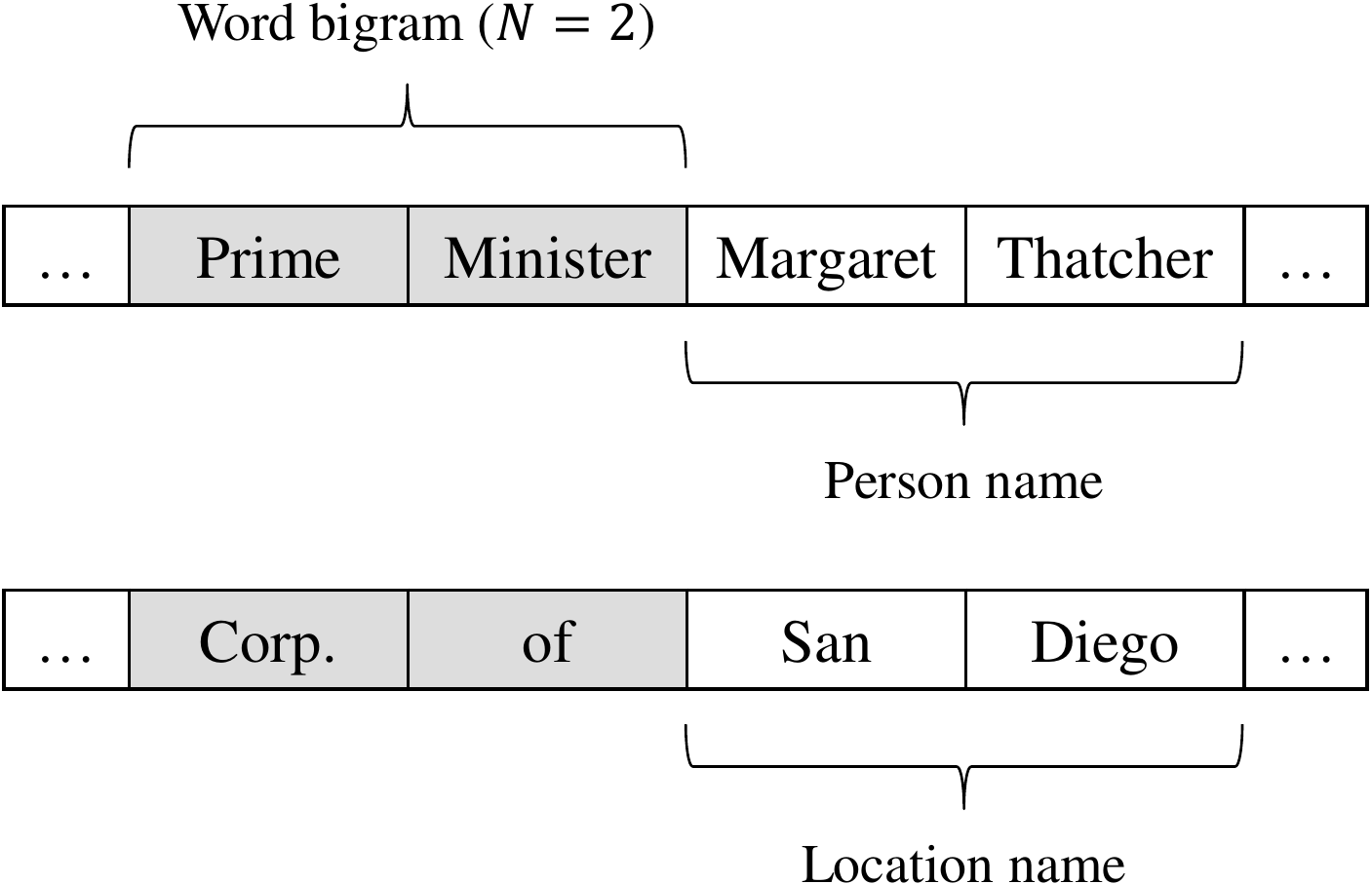}
  \caption{
  Examples of two types of named entities (NEs) and their left bigrams.
  The PER context (left bigram) contains the noun representing a person ``Minister.''
  The LOC context contains the preposition ``of,'' which tends to appear in various contexts.
  }
  \label{fig:Example}
\end{figure}

In this section, we analyze the behavior of our method and verify its effectiveness in mitigating the low- and zero-frequency problems.
To do so, we evaluate a task involving the use of LR estimators to predict the left context of named entities (NEs) in datasets.
We use two types of NEs---names of people and locations---which are tagged PER and LOC, respectively, and adopt word $N$-grams as their contexts.
Fig.~\ref{fig:Example} shows examples of the two types of NEs and their left bigram ($N=2$).
The left $N$-grams of PER tend to contain specific words, e.g., honorific titles or nouns representing a person.
As such, the words are crucial to predicting the PER context.
By contrast, LOC $N$-grams tend to contain prepositions, which are contained in various contexts.
Therefore, unlike in the PER context, contextual sequences (here indicating $N$-grams) are crucial.
These NEs with different context features are helpful for behavioral analysis.
Furthermore, the $N$-grams to be predicted are uniquely determined, which facilitates objective evaluation.
Finally, the $N$-gram frequency distributions are long-tail distributions, which forces us to handle numerous rare or unobserved items.
Therefore, we must be aware that methods used to handle them will significantly affect the prediction performance, allowing the differences between our method and the comparison methods described in Sect.~\ref{sec:Comparison_Methods} to be observed clearly.

\subsection{Dataset Descriptions and Experimental Conditions}
\begin{table*}[!t]
\caption{\# of bigrams contained in each dataset.}
\renewcommand{\arraystretch}{1.3}
\label{tab:Datasets_2-gram}
\centering
	\begin{tabular}{ c  c  r  r  c  r  r  c  r  r } \hline
		 \multicolumn{1}{ c }{Size of Train} & \multicolumn{1}{ c }{\multirow{2}{*}{Dataset}} & \multicolumn{2}{ c }{$O_{\rm ALL}$} && \multicolumn{2}{ c }{$O_{\rm LOC}$} &&  \multicolumn{2}{ c }{$O_{\rm PER}$} \\ \cline{3-4} \cline{6-7} \cline{9-10}
		 \multicolumn{1}{ c }{$S_{\rm tr}$} & & \multicolumn{1}{ c }{types} & \multicolumn{1}{ c }{total} && \multicolumn{1}{ c }{types} & \multicolumn{1}{ c }{total} && \multicolumn{1}{ c }{types} & \multicolumn{1}{ c }{total} \\ \hline \hline
		\multirow{3}{*}{10,000} & Train & 1,468,292 & 4,002,930 && 31,294 & 64,072 && 41,123 & 67,780 \\
		& Valid & 230,528 & 401,445 && 4,318 & 6,116 && 5,795 & 7,463 \\
		& Test & 231,931 & 403,145 && 4,164 & 5,876 && 6,035 & 7,635 \\ \hline
		\multirow{3}{*}{2,500} & Train &  493,368 & 1,000,627 && 9,983 & 15,965 && 12,458 & 17,292 \\
		& Valid & 237,805 & 415,369 && 4,752 & 6,736 && 5,522 & 6,973 \\
		& Test & 233,143 & 402,320 && 4,448 & 6,372 && 5,318 & 6,622 \\ \hline
	\end{tabular}
\end{table*}

\begin{table*}[!t]
\caption{\# of 4-grams contained in each dataset.}
\renewcommand{\arraystretch}{1.3}
\label{tab:Datasets_4-gram}
\centering
	\begin{tabular}{ c  c  r  r  c  r  r  c  r  r } \hline
		 \multicolumn{1}{ c }{Size of Train} & \multicolumn{1}{ c }{\multirow{2}{*}{Dataset}} & \multicolumn{2}{ c }{$O_{\rm ALL}$} && \multicolumn{2}{ c }{$O_{\rm LOC}$} &&  \multicolumn{2}{ c }{$O_{\rm PER}$} \\ \cline{3-4} \cline{6-7} \cline{9-10}
		 \multicolumn{1}{ c }{$S_{\rm tr}$} & & \multicolumn{1}{ c }{types} & \multicolumn{1}{ c }{total} && \multicolumn{1}{ c }{types} & \multicolumn{1}{ c }{total} && \multicolumn{1}{ c }{types} & \multicolumn{1}{ c }{total} \\ \hline \hline
		\multirow{3}{*}{10,000} & Train & 3,618,822 & 3,982,930 && 59,167 & 63,665 && 64,729 & 67,437 \\
		& Valid & 385,604 & 399,445 && 5,931 & 6,064 && 7,355 & 7,430 \\
		& Test & 388,401 & 401,145 && 5,713 & 5,832 && 7,497 & 7,593 \\ \hline
		\multirow{3}{*}{2,500} & Train & 947,710 & 995,627 && 15,305 & 15,851 && 16,919 & 17,219 \\
		& Valid & 400,336 & 413,369 && 6,574 & 6,699 && 6,861 & 6,940 \\
		& Test & 387,489 & 400,320 && 6,212 & 6,327 && 6,537 & 6,589 \\ \hline
	\end{tabular}
\end{table*}

We used NE-tagged datasets created by applying the following steps.
First, we classified and tagged entities in the 1987 Wall Street Journal corpus\footnote{\url{https://catalog.ldc.upenn.edu/LDC2000T43}} using the Stanford Named Entity Recognizer~\cite{Finkel:05}.
We then randomly sampled articles from the tagged corpus and allocated them to the training, validation, and test sets.
The validation and test set sizes were fixed at 1,000 articles in all cases, while the training set size $S_{\rm tr}$ was changed according to the following conditions, all of which had two choices\footnote{
We conducted experiments with $N$ varied from two to four.
We obtained similar results for the $N=2$ and $N=3$ cases.
Owing to the large degree of sparseness, we expected that the results for $N>4$ would not differ significantly from those of the $N=4$ cases;
therefore, we set $N=2$ and 4 as the $N$-gram order choices.}.
\begin{itemize}
  \setlength{\leftskip}{-0.4cm}
  \item $N$-gram order $N$: 2, 4
  \item training set size $S_{\rm tr}$: 10,000 articles, 2,500 articles
  \item NE type: location names (LOC), person names (PER)
\end{itemize}
We then conducted experiments at eight condition settings.
The first two conditions were for controlling training set sparseness; the last was for predicting different types of context.
The dataset descriptions are shown in Tables~\ref{tab:Datasets_2-gram} and \ref{tab:Datasets_4-gram}. 
Note that articles in the validation and test sets were resampled for different training set sizes.
$O_{\rm ALL}$ indicates that $N$-gram occurrences are at any position in a dataset.
$O_{\rm LOC}$ and $O_{\rm PER}$ indicate occurrences at the left side of LOC and PER, respectively, in a dataset.

\subsection{Experimental Procedure}

For any $N$-gram $w$ in the training set, we counted the frequencies on both the left side of the NEs and at any position.
Similarly, we counted the frequencies of $\{\Psi_kw\}_{k=1}^N$ to estimate $r_{\rm item}(w)$.
We then estimated $r(w)$ using the frequencies for each $w$ in the test set as follows:
\begin{align*}
r(w) = \frac{p_{\rm NE}(w)}{p_{\rm ALL}(w)},
\end{align*}
where $p_{\rm NE}(w)$ and $p_{\rm ALL}(w)$ are the occurrence probabilities of $w$ on the left side of the NEs and at any position in the training set, respectively.

We sorted and ranked $w$ in descending order of their estimates and evaluated the top 8,000.
If $w$ occurred on the left side of the NEs more than once in the test set, it was labeled ``True''; otherwise, it was labeled ``False.''
Finally, we constructed a rank--recall curve on a graph in which the horizontal and vertical axes represent rank and recall, respectively.
From the curve, we can see the approximate precision.
The slope of a straight line connecting the origin to a point on the curve is proportional to the precision at that point.
The recall and precision are defined as follows:
\begin{align*}
{\rm recall} = \frac{|\{w | w \in R\}|}{|R|}, \quad {\rm precision} = \frac{|\{w | w \in R\}|}{|\{w\}|},
\end{align*}
where $w$ is the $N$-gram to be evaluated and $R$ is the set of $N$-grams occurring on the left side of the NEs in the test set, i.e., the correction set.

\subsection{Comparison Methods}\label{sec:Comparison_Methods}

In this section, we describe the comparison methods and the parameter tuning process.
We used the following comparison methods, which have different behaviors for low-frequency (rare) or zero-frequency (unobserved) elements or for both types of elements.

A baseline (B for short, Eq.(\ref{eq:Indirect_MLE})) first obtains an MLE of each probability distribution and then takes the ratio of the MLEs.
This simple approach is often called ``indirect estimation'' because it involves probability estimation; however, it faces both the low- and zero-frequency problems.

Kikuchi:19 (K for short, Eq.(\ref{eq:solution_Kikuchi19})) is a variant of uLSIF that is applied to discrete distributions~\cite{Kikuchi:19}.
K introduces a regularization parameter $\lambda$ to mitigate the low-frequency problem but does not use frequencies of item units\footnote{For details of K and ITEM, see Sects.~\ref{sec:Kikuchi19} and \ref{sec:Itemized_LR}, respectively.}.
Thus, it cannot estimate LRs from unobserved elements.

In probability estimation, Kneser--Ney smoothing~\cite{Kneser:95} is widely used to solve the zero-frequency problem.
This smoothing uses lower order $N$-gram frequencies in addition to original $N$-gram frequencies.
A possible way to estimate $r(w)$ even from unobserved elements might be to smooth each probability estimate and then take the ratio of the estimates.
Therefore, we add the following method, which we call the Kneser--Ney-based method (KN for short), for comparison in $N=4$ settings:
\begin{align*}
 \widehat{r}_{\rm KN}(w) =& \frac{\dot{p}_{\rm nu}(w)}{\dot{p}_{\rm de}(w)}, \\
 \dot{p}_*(w) =& p_*^{\rm S}(a_1)p_*^{\rm S}(a_2 \mid a_1) \\
 & p_*^{\rm S}(a_3 \mid a_1a_2)p_*^{\rm S}(a_4 \mid a_1a_2a_3),
\end{align*}
where $\dot{p}_{\rm de}(w)$ and $\dot{p}_{\rm nu}(w)$ are modeled with smoothed conditional probabilities for $\{a_k\}_{k=1}^4$, and $p_*^{\rm S}$ is the (interpolated) Kneser--Ney estimator and has a discount parameter $d$.
That is to say, KN has eight parameters in total.
Note that $* \in \{\rm de, nu\}$ is an index that distinguishes the denominator and numerator of a likelihood ratio.

Itemized LR (ITEM for short, E.q.(\ref{eq:itemized_estimated})) mitigates the zero-frequency problem by itemizing $N$-grams to item units\footnotemark[1].
The low-frequency problem can be mitigated by using K for the LR estimation of each unit.
As this approach does not use original $N$-gram frequencies, the itemization sacrifices dependencies between items.
This is equivalent to our model without the dependency term $t_{\rm d}(w)$, and we therefore include it in our comparison to confirm the effectiveness of $t_{\rm d}(w)$.
To avoid duplication in the model's notation, the regularization parameter is replaced by $\lambda'$.

With the exception of B, all of the evaluated methods have tuning parameters.
We therefore regarded the validation sets as test sets and constructed rank--recall curves with the regularization parameters changed over the range $10^{-9}, 10^{-8}, \ldots ,10^{-1}$.
Finally, we chose the optimal value that maximized the area under the rank--recall curve.
As our method uses two parameters---$\lambda_{\rm item}$ and $\lambda_{\rm d}$---the optimal pair was found by changing them simultaneously.
The chosen regularization parameters are listed in Tables~\ref{tab:Parameters_2-gram} and \ref{tab:Parameters_4-gram}.
As described above, KN has eight tuning parameters.
For each Kneser--Ney estimator, we varied the parameter $d$ over the range $0.1,0.2, \ldots ,0.9$ and chose the optimal value that maximized its likelihood.
This is a natural way to set $d$s when using the smoothing.

\begin{table}[!t]
\caption{Regularization parameters chosen for $N=2$.}
\renewcommand{\arraystretch}{1.3}
\label{tab:Parameters_2-gram}
\centering
	\begin{tabular}{ c  c  r  c  r  c  r  r } \hline
		\multicolumn{1}{ c }{Size of Train} & \multirow{2}{*}{NE} & \multicolumn{1}{ c }{K} && \multicolumn{1}{ c }{ITEM} && \multicolumn{2}{ c }{Ours} \\ \cline{3-3} \cline{5-5} \cline{7-8}
		\multicolumn{1}{ c }{$S_{\rm tr}$} && \multicolumn{1}{ c }{$\lambda$} && \multicolumn{1}{ c }{$\lambda'$} && \multicolumn{1}{ c }{$\lambda_{\rm item}$} & \multicolumn{1}{ c }{$\lambda_{\rm d}$} \\ \hline \hline
		\multirow{2}{*}{10,000} & LOC & $10^{-2}$ && $10^{-5}$ && $10^{-1}$ & $10^{-4}$ \\
		& PER & $10^{-3}$ && $10^{-5}$ && $10^{-4}$ & $10^{-5}$ \\ \hline
		\multirow{2}{*}{2,500} & LOC & $10^{-1}$ && $10^{-4}$ && $10^{-1}$ & $10^{-3}$ \\
		& PER & $10^{-1}$ && $10^{-5}$ && $10^{-4}$ & $10^{-5}$ \\ \hline
	\end{tabular}
\end{table}

\begin{table}[!t]
\caption{Regularization parameters chosen for $N=4$.}
\renewcommand{\arraystretch}{1.3}
\label{tab:Parameters_4-gram}
\centering
	\begin{tabular}{ c  c  r  c  r  c  r  r } \hline
		\multicolumn{1}{ c }{Size of Train} & \multirow{2}{*}{NE} & \multicolumn{1}{ c }{K} && \multicolumn{1}{ c }{ITEM} && \multicolumn{2}{ c }{Ours} \\ \cline{3-3} \cline{5-5} \cline{7-8}
		\multicolumn{1}{ c }{$S_{\rm tr}$} && \multicolumn{1}{ c }{$\lambda$} && \multicolumn{1}{ c }{$\lambda'$} && \multicolumn{1}{ c }{$\lambda_{\rm item}$} & \multicolumn{1}{ c }{$\lambda_{\rm d}$} \\ \hline \hline
		\multirow{2}{*}{10,000} & LOC & $10^{-6}$ && $10^{-6}$ && $10^{-5}$ & $10^{-8}$ \\
		& PER & $10^{-6}$ && $10^{-5}$ && $10^{-5}$ & $10^{-6}$ \\ \hline
		\multirow{2}{*}{2,500} & LOC & $10^{-1}$ && $10^{-5}$ && $10^{-5}$ & $10^{-6}$ \\
		& PER & $10^{-1}$ && $10^{-5}$ && $10^{-5}$ & $10^{-5}$ \\ \hline
	\end{tabular}
\end{table}

\subsection{Results}

We first show the respective rank--recall curves, which are summarized for different $N$.
Then, for behavioral analysis, we show some ranked $N$-grams.
Owing to space limitations, only the $N$-grams for $N=2$, $S_{\rm tr}=10,000$ and $N=4$, $S_{\rm tr}=2,500$ are shown.

\subsection*{Rank-Recall Curves for $N=2$}
\begin{figure}[!t]
  \begin{minipage}[b]{0.5\linewidth}
    \centering
    \includegraphics[keepaspectratio, scale=0.3]{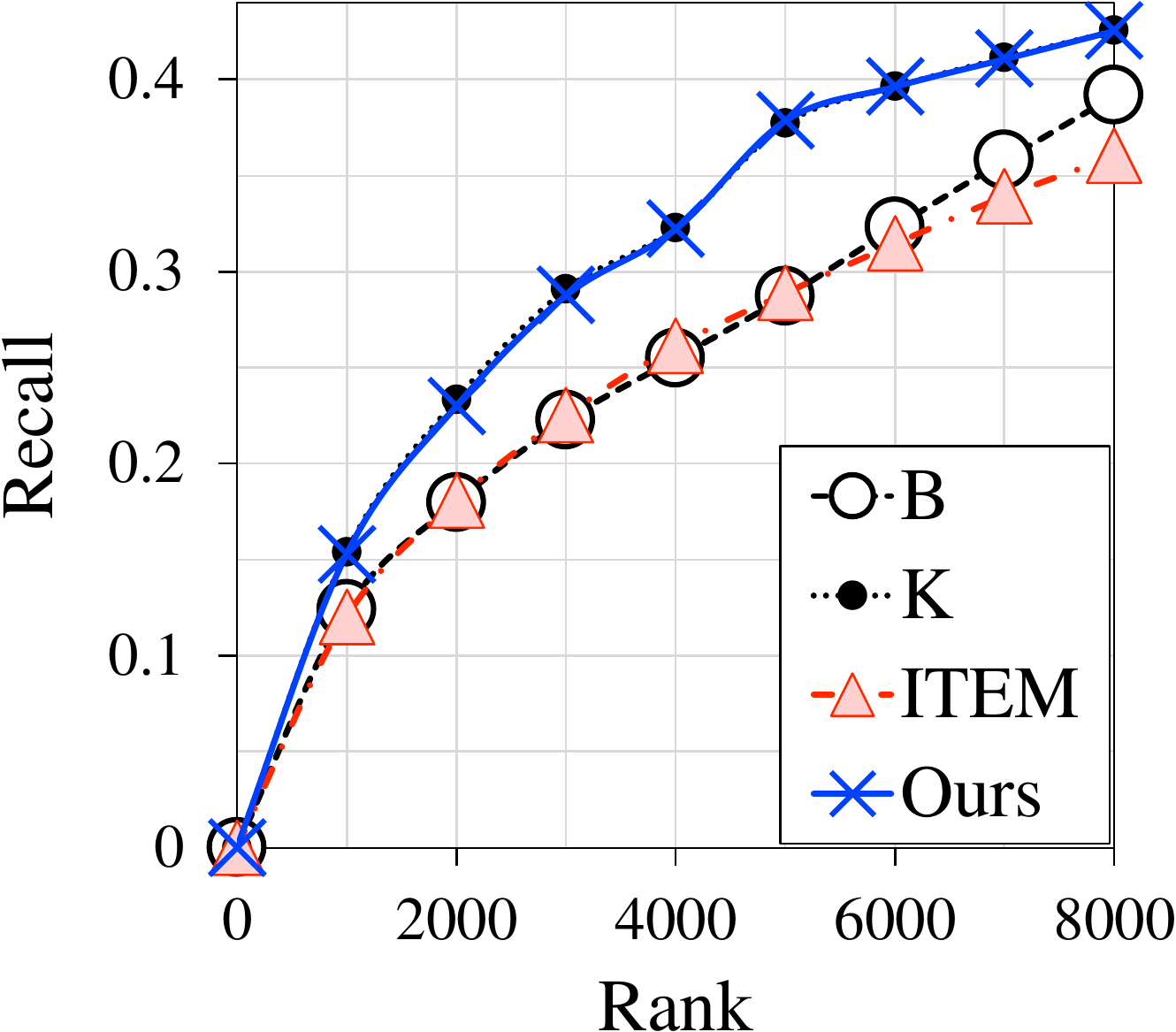}
    \subcaption{LOC, $S_{\rm tr}=10,000$}\label{fig:LOCATION_10000_2-gram}
  \end{minipage}
  \begin{minipage}[b]{0.5\linewidth}
    \centering
    \includegraphics[keepaspectratio, scale=0.3]{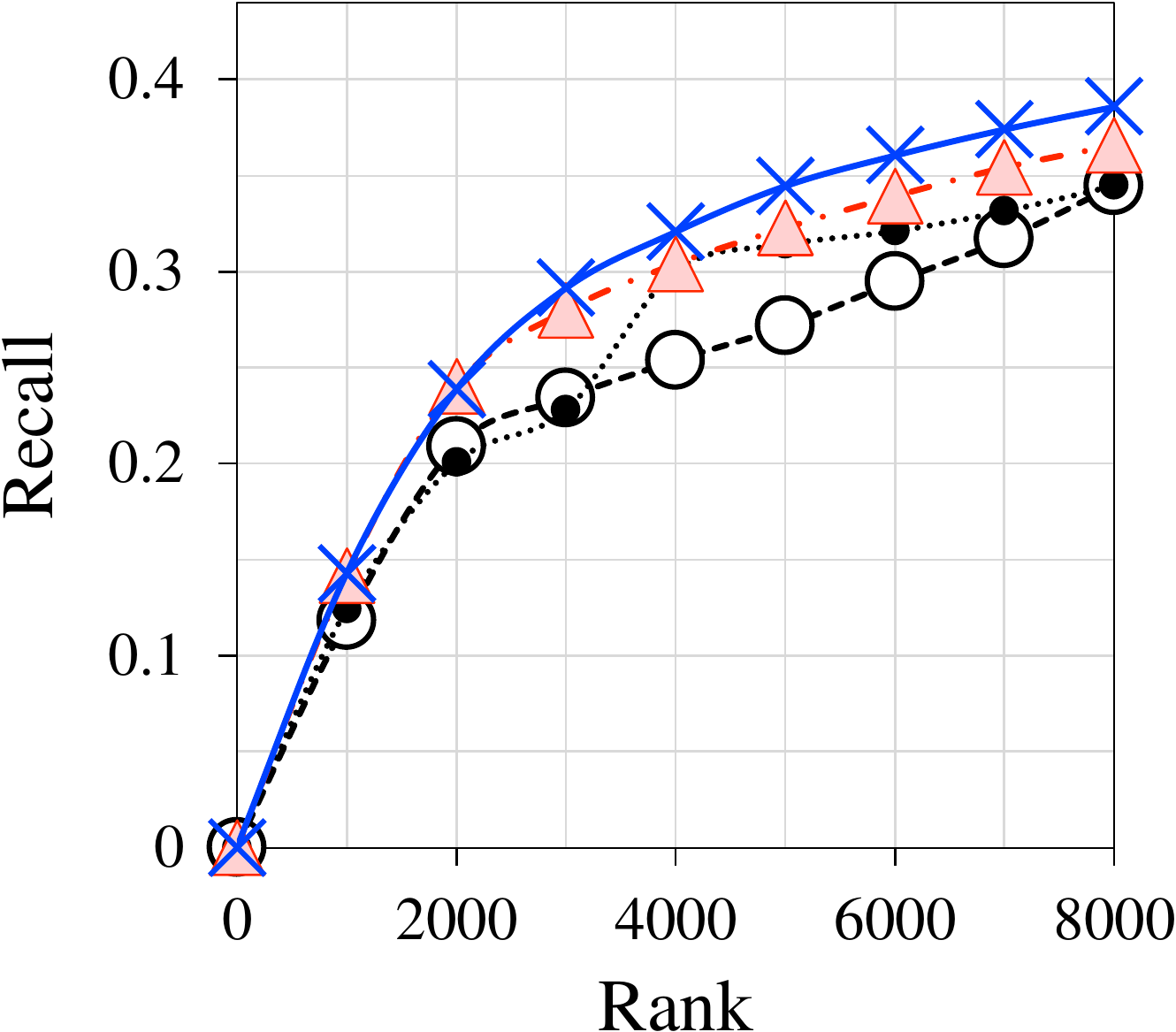}
    \subcaption{PER, $S_{\rm tr}=10,000$}\label{fig:PERSON_10000_2-gram}
  \end{minipage} \vspace{1pt} \\
  \begin{minipage}[b]{0.5\linewidth}
    \centering
    \includegraphics[keepaspectratio, scale=0.3]{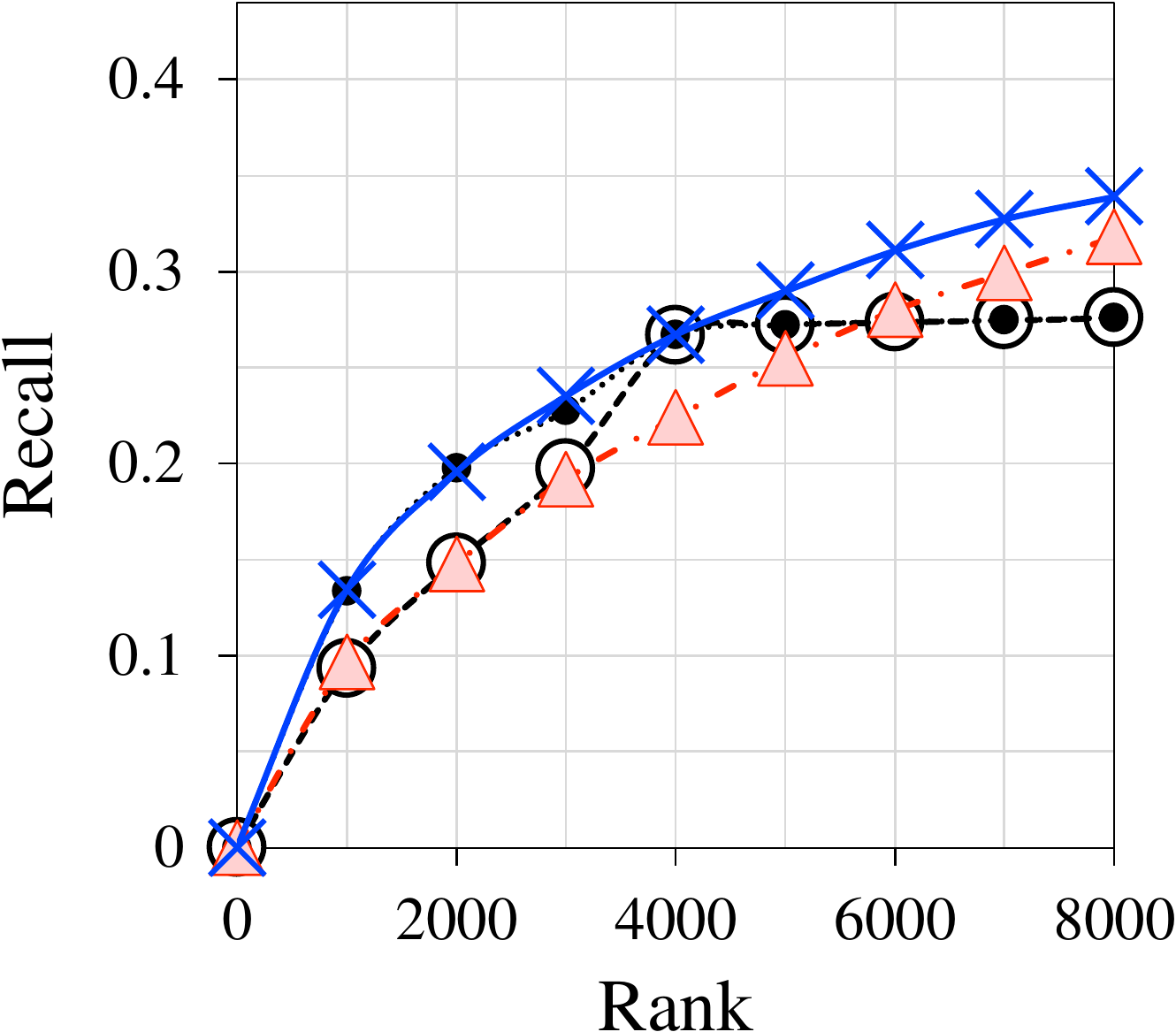}
    \subcaption{LOC, $S_{\rm tr}=2,500$}\label{fig:LOCATION_2500_2-gram}
  \end{minipage}
  \begin{minipage}[b]{0.5\linewidth}
    \centering
    \includegraphics[keepaspectratio, scale=0.3]{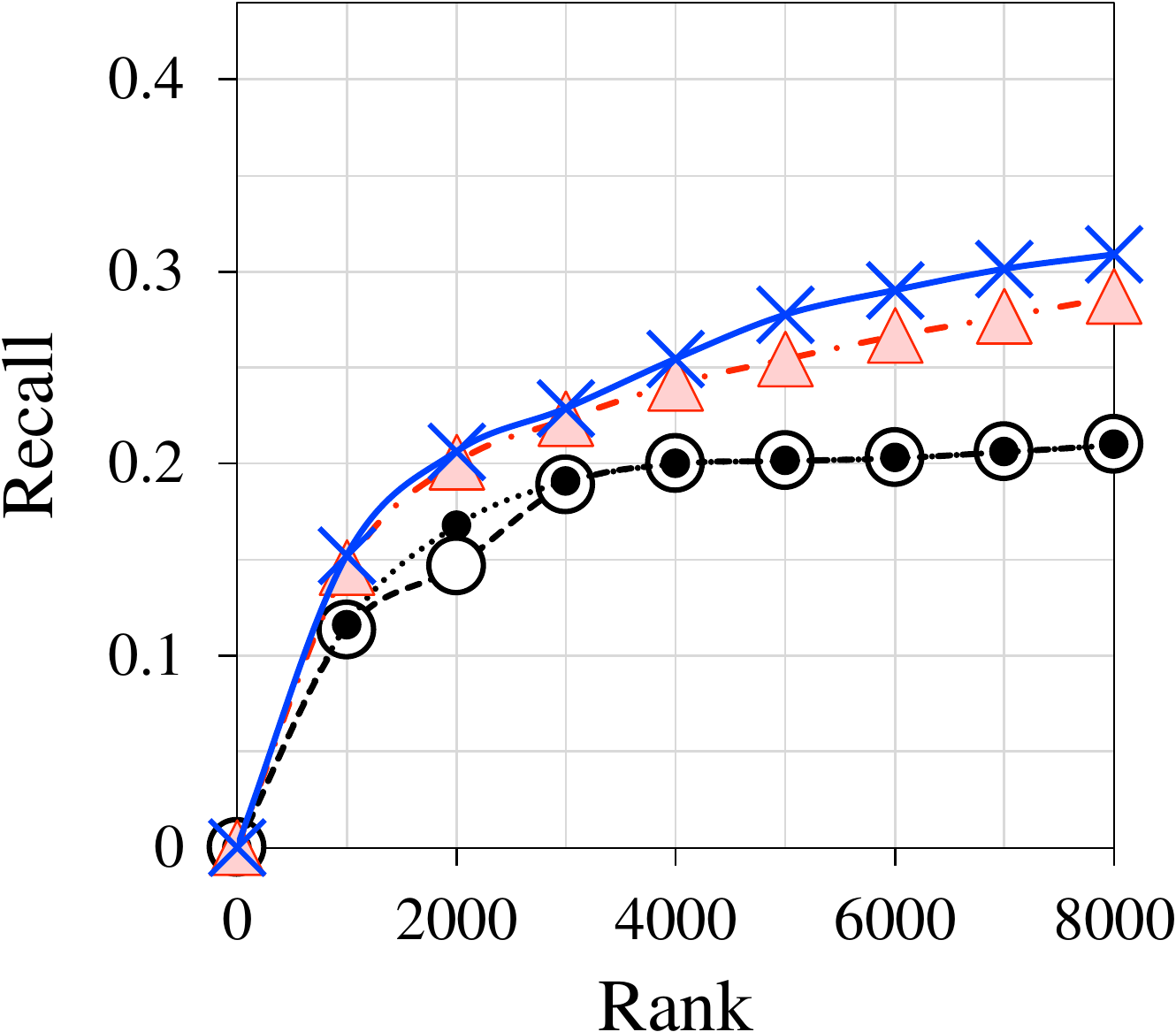}
    \subcaption{PER, $S_{\rm tr}=2,500$}\label{fig:PERSON_2500_2-gram}
  \end{minipage}
  \caption{
  Rank--recall curves for $N=2$.
  PER and LOC indicate that prediction targets are the left PER and LOC contexts, respectively.
  $S_{\rm tr}$ is the training set size.
  The horizontal and vertical axes of curves represent rank and recall, respectively.
  The slope of a straight line connecting the origin to a point on the curve is proportional to the precision at that point.
  For a given rank, the method with the largest value on the vertical axis, i.e., the highest recall, has the best performance.
  }
  \label{fig:Results_2-gram}
\end{figure}

Fig.~\ref{fig:Results_2-gram} shows the rank--recall curves for bigrams ($N=2$).
From this figure, we can see the following:

\begin{itemize}
  \item
  The baseline (B) performed worse than our method as a result of the low-frequency problem.
  In addition, Kikuchi:19 (K) performed generally better than B for $S_{\rm tr}=10,000$ because of its regularization.
  In particular, K was competitive with our method (Fig.~\ref{fig:Results_2-gram}(a)).
  However, both methods suffered from unobserved bigrams for $S_{\rm tr}=2,500$ and could not improve recall in the low-rank range (Figs.~\ref{fig:Results_2-gram}(c) and \ref{fig:Results_2-gram}(d)).
  \item
  Itemized LR (ITEM) demonstrated improving recall up to rank 8,000.
  This suggests that itemizing $N$-grams is effective in mitigating the zero-frequency problem.
  Moreover, this method produced the second-best performance results in PER context prediction, as shown in Figs.~\ref{fig:Results_2-gram}(b) and \ref{fig:Results_2-gram}(d).
  However, in terms of LOC context prediction, ITEM performed worse than our method and by far the worst for $S_{\rm tr}=10,000$ (Fig.~\ref{fig:Results_2-gram}(a)) despite having the richest training context.
  \item
  Our method, ITEM with $t_{\rm d}(w)$, was able to mitigate the zero-frequency problem while maintaining dependency between items.
  Moreover, its regularization mitigated the low-frequency problem.
  Overall, our method outperformed the other methods on all graphs excluding Fig.~\ref{fig:Results_2-gram}(a), in which it was competitive with K.
\end{itemize}

\subsection*{Ranked Bigram Examples for $S_{\rm tr}= 10,000$}
\begin{table*}[!t]
\caption{Ranked bigrams $w=a_1a_2$ for $S_{\rm tr}=10,000$.}
\renewcommand{\arraystretch}{1.3}
\label{tab:Bigrams_10000}
\centering
	\begin{tabular}{ c  l  r  r  c  r  r  r  r  c  r  r } \hline
		\multicolumn{1}{ c }{\multirow{2}{*}{$w$}} & \multicolumn{1}{ c }{\multirow{2}{*}{$a_1a_2$}} & \multicolumn{2}{ c }{\#$w$} && \multicolumn{4}{ c }{Rank} & \multirow{2}{*}{T/F}  & \multicolumn{2}{ c }{$\widehat{r}_{\rm ours}(w)$} \\ \cline{3-4} \cline{6-9} \cline{11-12}
		& & \multicolumn{1}{ c }{$O_{\rm ALL}$} & \multicolumn{1}{ c }{$O_{\rm LOC}$} && \multicolumn{1}{ c }{B} & \multicolumn{1}{ c }{K} & \multicolumn{1}{ c }{ITEM} & \multicolumn{1}{ c }{Ours} &&  \multicolumn{1}{ c }{$\widehat{r}_{\rm item}(w)$} &  \multicolumn{1}{ c }{$\widehat{t}_{\rm d}(w)$} \\ \hline \hline
		$w_{\rm A}$ & Hospital in & 22 & 22 && 1 & 190 & 86 & 115 & T & $4.88 \times 10^{-3}$ & $3.25$ \\
		$w_{\rm B}$ & court in & 186 & 156 && 602 & 16 & 180 & 3 & T & $3.40 \times 10^{-2}$ & $16.61$ \\
		$w_{\rm C}$ & chair of & 1 & 1 && 1 & 4,153 & -- & 4,119 & F & $9.94 \times 10^{-5}$ & $0.16$ \\
		$w_{\rm D}$ & Europe in & 12 & 0 && -- & -- & 147 & -- & F & $9.54 \times 10^{-3}$ & $-2.78 \times 10^{-4}$ \\ \hline
		\multicolumn{1}{ c }{\multirow{2}{*}{$w$}} & \multicolumn{1}{ c }{\multirow{2}{*}{$a_1a_2$}} & \multicolumn{2}{ c }{\#$w$} && \multicolumn{4}{ c }{Rank} & \multirow{2}{*}{T/F}  & \multicolumn{2}{ c }{$\widehat{r}_{\rm ours}(w)$} \\ \cline{3-4} \cline{6-9} \cline{11-12}
		& & \multicolumn{1}{ c }{$O_{\rm ALL}$} & \multicolumn{1}{ c }{$O_{\rm PER}$} && \multicolumn{1}{ c }{B} & \multicolumn{1}{ c }{K} & \multicolumn{1}{ c }{ITEM} & \multicolumn{1}{ c }{Ours} &&  \multicolumn{1}{ c }{$\widehat{r}_{\rm item}(w)$} &  \multicolumn{1}{ c }{$\widehat{t}_{\rm d}(w)$} \\ \hline \hline
		$w_{\rm E}$ & Prime Minister & 228 & 228 && 1 & 11 & 1 & 19 & T & $614.42$ & $-472.42$ \\
		$w_{\rm F}$ & says Mr. & 632 & 537 && 1,210 & 2 & 89 & 159 & T & $287.75$ & $-223.42$ \\
		$w_{\rm G}$ & \$53.9 million, & 1 & 1 && 1 & 3,045 & -- & 6,069 & F & $2.18 \times 10^{-2}$ & $1.44$ \\
		$w_{\rm H}$ & So President & 0 & 0 && -- & -- & 260 & 82 & T & $81.09$ & $0$ \\ \hline
	\end{tabular}
\end{table*}
\begin{table}[tb]
\small
\caption{Itemized frequencies of unobserved bigrams for $S_{\rm tr}=10,000$.}
\renewcommand{\arraystretch}{1.3}
\label{tab:Itemized_10000}
\centering
	\begin{tabular}{ c  l  r  r  c  r  r } \hline
		\multicolumn{1}{ c }{\multirow{2}{*}{$w$}} & \multicolumn{1}{ c }{\multirow{2}{*}{$a_1a_2$}} & \multicolumn{2}{ c }{\# $(a_1 \  \bullet)$} && \multicolumn{2}{ c }{\# $(\bullet \  a_2)$} \\ \cline{3-4} \cline{6-7}
		& & \multicolumn{1}{ c }{$O_{\rm ALL}$} & \multicolumn{1}{ c }{$O_{\rm LOC}$} && \multicolumn{1}{ c }{$O_{\rm ALL}$} & \multicolumn{1}{ c }{$O_{\rm LOC}$} \\ \hline \hline
		$w_{\rm D}$ & Europe in & 325 & 45 && 73,816 & 10,312 \\ \hline
		\multicolumn{1}{ c }{\multirow{2}{*}{$w$}} & \multicolumn{1}{ c }{\multirow{2}{*}{$a_1a_2$}} & \multicolumn{2}{ c }{\# $(a_1 \  \bullet)$} && \multicolumn{2}{ c }{\# $(\bullet \ a_2)$} \\ \cline{3-4} \cline{6-7}
		& & \multicolumn{1}{ c }{$O_{\rm ALL}$} & \multicolumn{1}{ c }{$O_{\rm PER}$} && \multicolumn{1}{ c }{$O_{\rm ALL}$} & \multicolumn{1}{ c }{$O_{\rm PER}$} \\ \hline \hline
		$w_{\rm H}$ & So President & 559 & 32 && 1,214 & 1,172 \\ \hline
	\end{tabular}
\end{table}

To clarify the factors underlying differences in performance among the methods, we analyzed the behavior of each method qualitatively.
Table~\ref{tab:Bigrams_10000} shows word bigrams ranked by method for $S_{\rm tr}=10,000$.
In the table, ``--'' indicates that the bigram is ranked below the top 8,000;
``\#'' indicates the frequencies in the training set.
Table~\ref{tab:Itemized_10000} shows itemized frequencies of the unobserved bigrams in Table~\ref{tab:Bigrams_10000}.
The rank--recall curves corresponding to these tables are shown in Figs.~\ref{fig:Results_2-gram}(a) and \ref{fig:Results_2-gram}(b).
From the tables, we can see the following:

\begin{itemize}
  \item
  B ranked $w_{\rm A}$, $w_{\rm C}$, $w_{\rm E}$ and $w_{\rm G}$ at the top.
  These always occurred on the left side of NEs in the training set.
  While B allowed placing of the true bigrams $w_{\rm A}$ and $w_{\rm E}$ within high ranks, the rare and false bigrams $w_{\rm C}$ and $w_{\rm G}$, respectively, were also placed within high ranks because it overestimated the LRs of rare bigrams.
  The frequent bigrams $w_{\rm B}$ and $w_{\rm F}$, which should have been in the higher ranks, were ranked lower than they were by the other methods.
  Therefore, B dose not have good performance for any $S_{\rm tr}$ or NE type.
  \item
  K placed frequent and true bigrams $w_{\rm A}$, $w_{\rm B}$, $w_{\rm E}$ and $w_{\rm F}$ within the higher ranks while placing the rare and false bigrams $w_{\rm C}$ and $w_{\rm G}$ in the lower ranks.
  This suggests that the regularization worked well and resulted in better performance than B.
  \item
  ITEM behaved in a manner similar to K except for the unobserved bigrams $w_{\rm D}$ and $w_{\rm H}$.
  This suggests that itemizing $w$ was not harmful for observed bigrams.
  However, this method could not take into account the dependency between $a_1$ and $a_2$ and therefore computed item pair estimates without considering their co-occurrence.
  Consequently, the false bigram $w_{\rm D}$, which is not a pair of co-occurring items but rather a pair of separate, frequent items, as shown in Table~\ref{tab:Itemized_10000}, was placed in a high rank.
  We can infer that this degraded the prediction performance of LOC context, which required a strong dependency.
  \item
  Our method can adjust the dependency strength according to usage and application.
  Because LOC context prediction requires a strong dependency,
  the parameters were chosen so that $\lambda_{\rm item} > \lambda_{\rm d}$, as shown in Table~\ref{tab:Parameters_2-gram}.
  Thus, $\widehat{t}_{\rm d}(w)$ was much larger than $\widehat{r}_{\rm item}(w)$ except for $w_{\rm D}$.
  Meanwhile, $\widehat{t}_{\rm d}(w_{\rm D})$ was smaller than $\widehat{r}_{\rm item}(w_{\rm D})$ owing to the low dependency; i.e., \#$w_{\rm D}$ of $O_{\rm LOC}$ was zero.
  As a result, $w_{\rm D}$ had the small estimate and was placed in a lower rank.
  In the PER context prediction process, by contrast, the parameters were chosen so that the difference between them was smaller than in the LOC case.
  Unlike LOC cases, $\widehat{r}_{\rm item}(w)$ was much larger than $\widehat{t}_{\rm d}(w)$ except for $w_{\rm G}$.
  In particular, $\widehat{r}_{\rm ours}(w_{\rm H})$ was large even though the \#$w_{\rm H}$ of $O_{\rm ALL}$ and $O_{\rm PER}$ were both zero.
  ($\bullet$ President) was the most frequent and crucial unit, as shown in Table~\ref{tab:Itemized_10000}.
  Here, our method paid attention to occurring individual items and not their dependencies.
  As a result, $w_{\rm H}$ had the large estimate and was placed in a high rank.
  Overall, we can hypothesize that this behavior resulted in the best performance for predictions in different contexts.
\end{itemize}

\subsection*{Rank-Recall Curves for $N=4$}
\begin{figure}[!t]
  \begin{minipage}[b]{0.5\linewidth}
    \centering
    \includegraphics[keepaspectratio, scale=0.3]{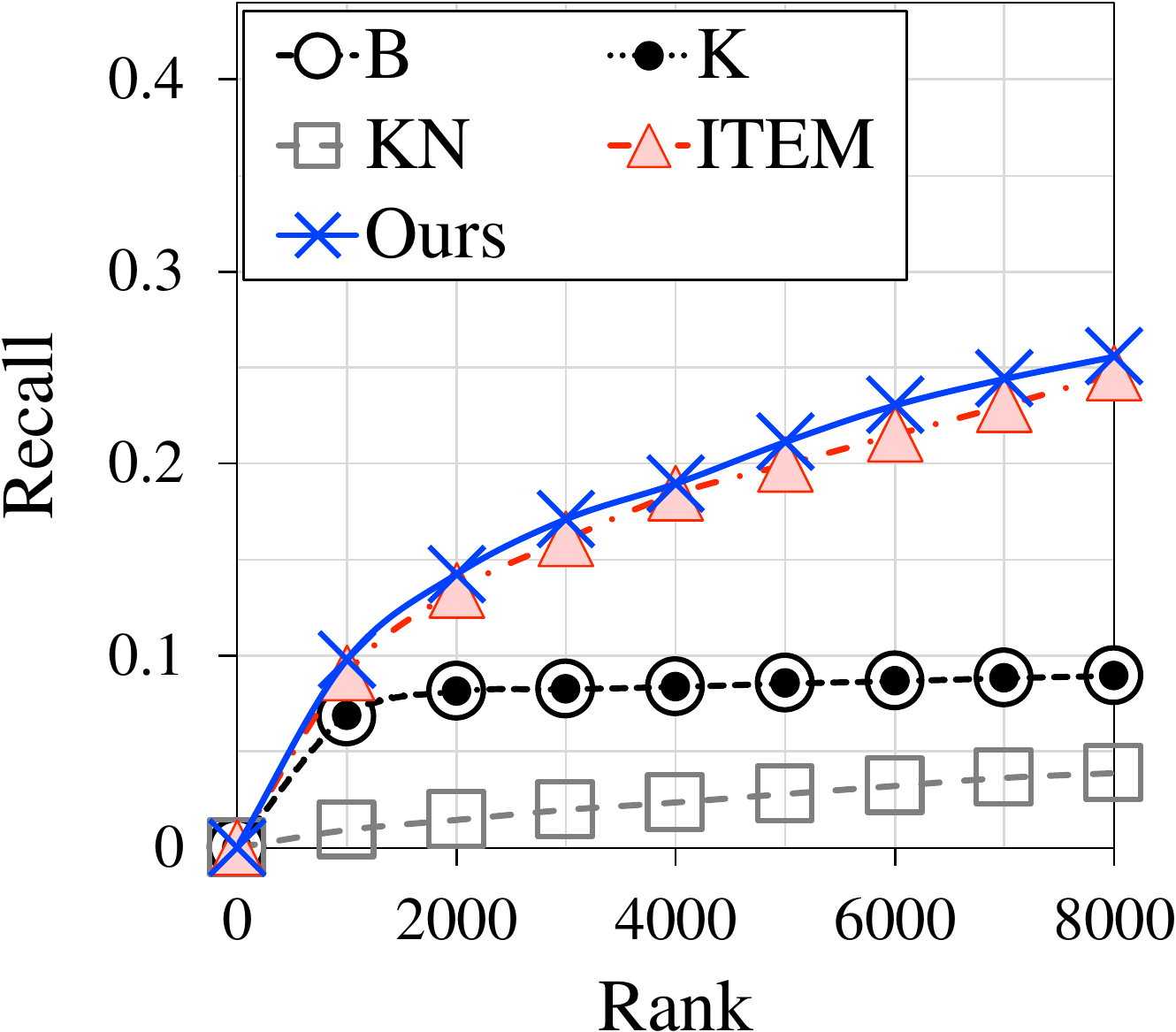}
    \subcaption{LOC, $S_{\rm tr}=10,000$}\label{fig:LOCATION_10000_4-gram}
  \end{minipage}
  \begin{minipage}[b]{0.5\linewidth}
    \centering
    \includegraphics[keepaspectratio, scale=0.3]{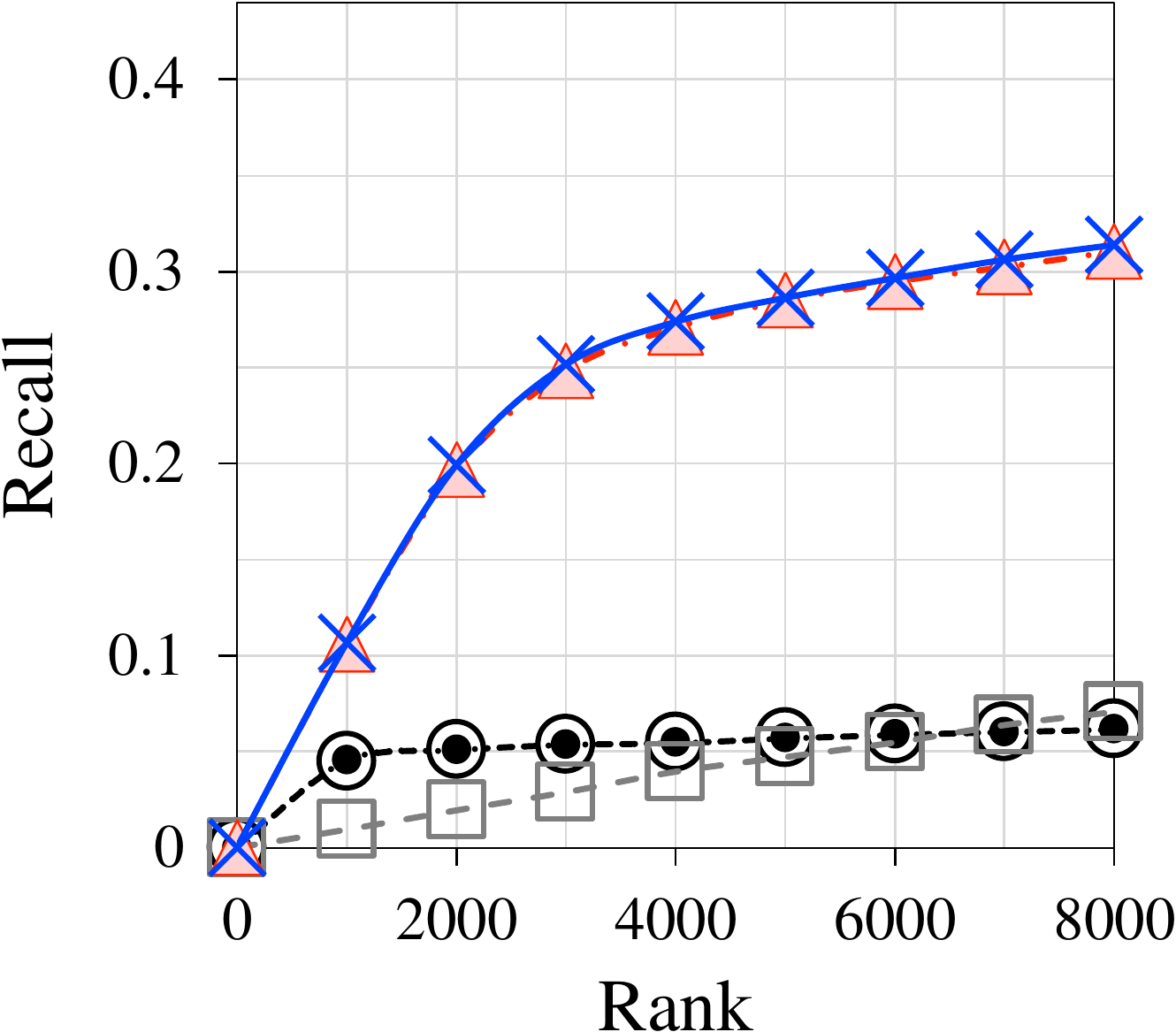}
    \subcaption{PER, $S_{\rm tr}=10,000$}\label{fig:PERSON_10000_4-gram}
  \end{minipage} \vspace{1pt} \\
  \begin{minipage}[b]{0.5\linewidth}
    \centering
    \includegraphics[keepaspectratio, scale=0.3]{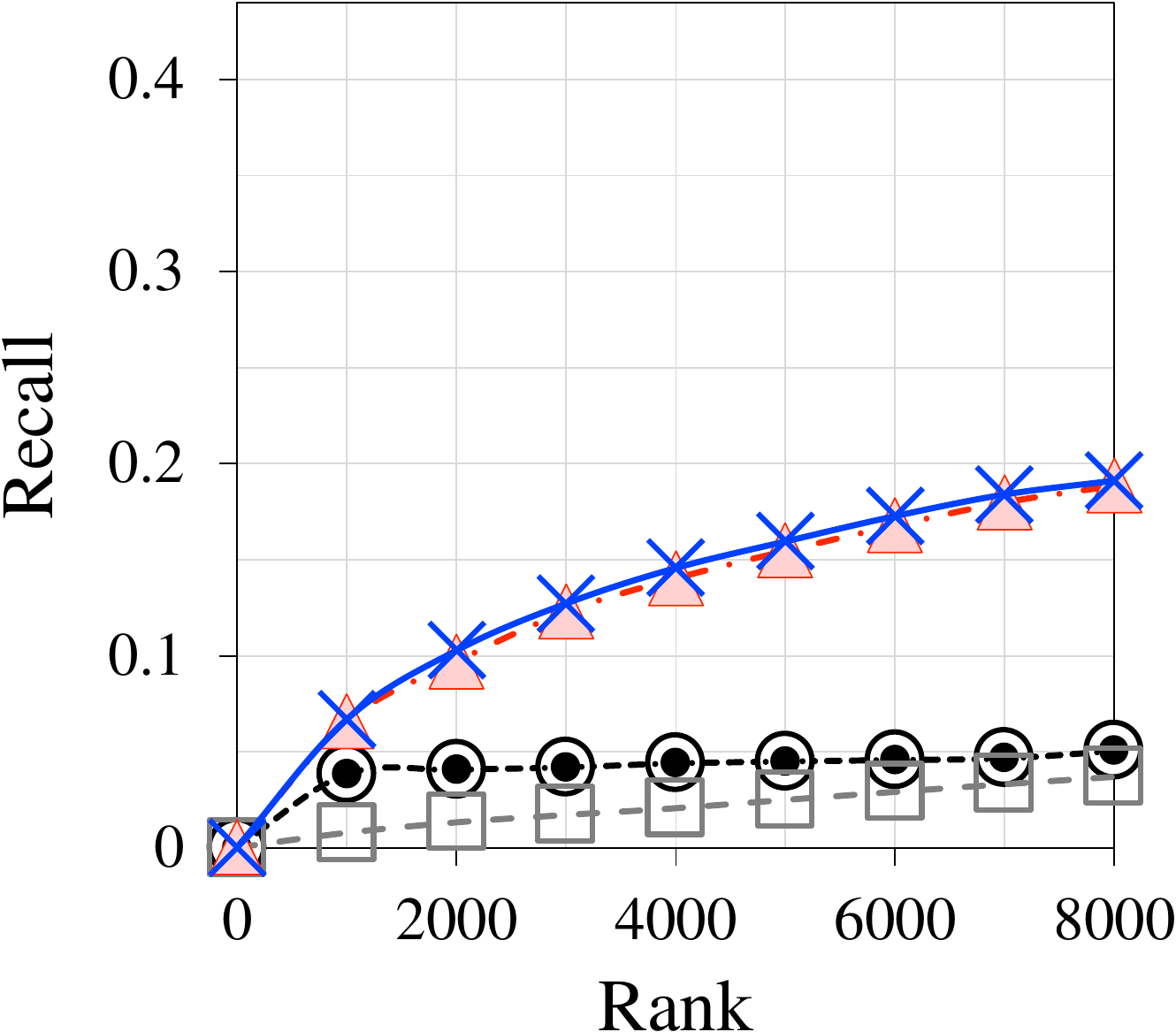}
    \subcaption{LOC, $S_{\rm tr}=2,500$}\label{fig:LOCATION_2500_4-gram}
  \end{minipage}
  \begin{minipage}[b]{0.5\linewidth}
    \centering
    \includegraphics[keepaspectratio, scale=0.3]{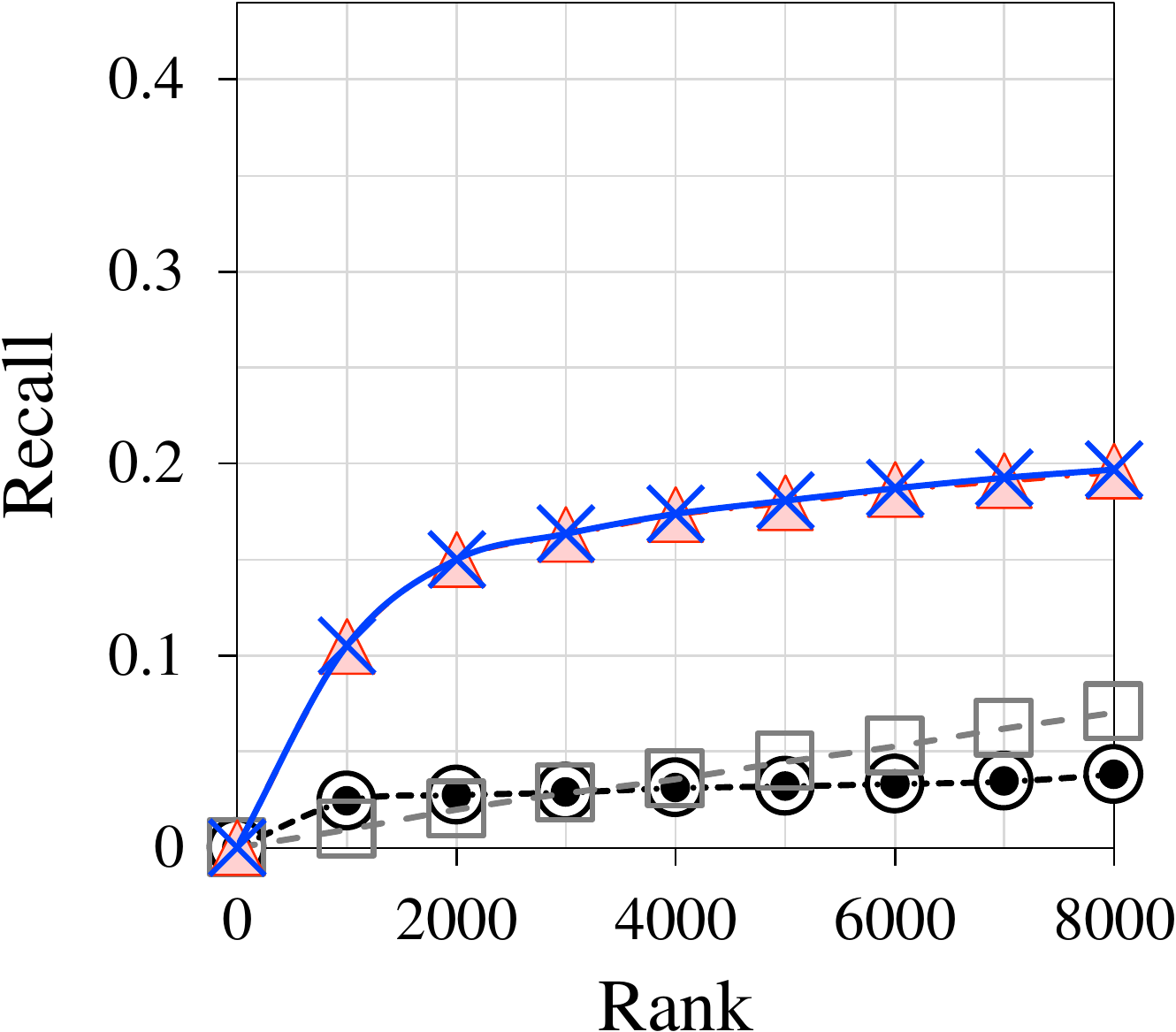}
    \subcaption{PER, $S_{\rm tr}=2,500$}\label{fig:PERSON_2500_4-gram}
  \end{minipage}
  \caption{
  Rank--recall curves for $N=4$.
  PER and LOC indicate that prediction targets are left PER and LOC contexts, respectively.
  $S_{\rm tr}$ is the training set size.
  The horizontal and vertical axes of the curves represent rank and recall, respectively.
  The slope of a straight line connecting the origin to a point on the curve is proportional to the precision at that point.
  For a given rank, the method with the largest value on the vertical axis, i.e., the highest recall, has the best performance.
  }
  \label{fig:Results_4-gram}
\end{figure}

Fig.~\ref{fig:Results_4-gram} shows the rank--recall curves for four-grams ($N=4$).
In this case, the context was much sparser than for $N=2$, and therefore all methods suffered from a lack of frequency.
From the figure, we can see the following:

\begin{itemize}
  \item
  B and K performed in a similarly poor manner in all rank--recall curves. 
  Their precisions improved slightly in the high-rank ranges, but did not improve thereafter even at $S_{\rm tr}=10,000$ cases.
  This suggests that the methods discovered very few contexts in the sparse data settings.
  \item
 The Kneser--Ney-based method (KN) also performed poorly.
  In probability estimation, Kneser--Ney smoothing is known to be an effective method for solving the zero-frequency problem.
  However, we found that taking the ratio of smoothed estimates significantly degraded performance.
  \item
  In the high-rank ranges of all curves, ITEM and our method performed with the highest precisions.
  Furthermore, they continued to improve their recall up to rank 8,000, a result that confirms the effectiveness of itemizing $N$-grams.
  Unlike the bigram cases, these two methods performed similarly; we can infer from this that the effect of $t_{\rm d}(w)$ was weakened.
  However, our method performed slightly better than ITEM in LOC context prediction, as shown in Figs.~\ref{fig:Results_4-gram}(a) and \ref{fig:Results_4-gram}(c).
  This result confirms that incorporating the dependency is also effective for large-$N$ settings.
\end{itemize}

\subsection*{Ranked Four-gram Examples for $S_{\rm tr}= 2,500$}
\begin{table*}[!t]
\caption{Ranked four-grams $w=a_1a_2a_3a_4$ for $S_{\rm tr}=2,500$.}
\renewcommand{\arraystretch}{1.3}
\label{tab:4-grams_2500}
\centering
	\begin{tabular}{ c  l  r  r  c  r  r  r  r  r  c  r  r } \hline
		\multicolumn{1}{ c }{\multirow{2}{*}{$w$}} & \multicolumn{1}{ c }{\multirow{2}{*}{$a_1a_2a_3a_4$}} & \multicolumn{2}{ c }{\#$w$} && \multicolumn{5}{ c }{Rank} & \multirow{2}{*}{T/F}  & \multicolumn{2}{ c }{$\widehat{r}_{\rm ours}(w)$} \\ \cline{3-4} \cline{6-10} \cline{12-13}
		& & \multicolumn{1}{ c }{$O_{\rm ALL}$} & \multicolumn{1}{ c }{$O_{\rm LOC}$} && \multicolumn{1}{ c }{B} & \multicolumn{1}{ c }{K} & \multicolumn{1}{ c }{KN} & \multicolumn{1}{ c }{ITEM} & \multicolumn{1}{ c }{Ours} && \multicolumn{1}{ c }{$\widehat{r}_{\rm item}(w)$} &  \multicolumn{1}{ c }{$\widehat{t}_{\rm d}(w)$} \\ \hline \hline
		$w_{\rm A}$ & University of California at & 7 & 7 && 1 & 8 & -- & 307 & 678 & T & $204.63$ & $-124.16$ \\
		$w_{\rm B}$ & a large number of & 5 & 1 && 487 & 518 & -- & -- & 4,850 & F & $5.07$ & $6.25$ \\
		$w_{\rm C}$ & in West Germany, France, & 0 & 0 && 611 & 611 & -- & 57 & 41 & T & $1655.59$ & $0$ \\
		$w_{\rm D}$ & Christ Church in Grosse & 0 & 0 && 611 & 611 & 169 & -- & -- & F & $0$ & $0$ \\ \hline
		\multicolumn{1}{ c }{\multirow{2}{*}{$w$}} & \multicolumn{1}{ c }{\multirow{2}{*}{$a_1a_2a_3a_4$}} & \multicolumn{2}{ c }{\#$w$} && \multicolumn{5}{ c }{Rank} & \multirow{2}{*}{T/F}  & \multicolumn{2}{ c }{$\widehat{r}_{\rm ours}(w)$} \\ \cline{3-4} \cline{6-10} \cline{12-13}
		& & \multicolumn{1}{ c }{$O_{\rm ALL}$} & \multicolumn{1}{ c }{$O_{\rm PER}$} && \multicolumn{1}{ c }{B} & \multicolumn{1}{ c }{K} & \multicolumn{1}{ c }{KN} & \multicolumn{1}{ c }{ITEM} & \multicolumn{1}{ c }{Ours} && \multicolumn{1}{ c }{$\widehat{r}_{\rm item}(w)$} &  \multicolumn{1}{ c }{$\widehat{t}_{\rm d}(w)$} \\ \hline \hline
		$w_{\rm E}$ & In an interview, Mr. & 7 & 7 && 1 & 4 & -- & 38 & 67 & T & $1560.35$ & $-620.29$ \\
		$w_{\rm F}$ & Procter \& Gamble Co. & 4 & 1 && 290 & 314 & -- & 3,122 & 2,799 & F & $7.60$ & $1.96$ \\
		$w_{\rm G}$ & promised by Prime Minister & 0 & 0 && 415 & 415 & -- & 14 & 13 & T & $3153.59$ & $0$ \\
		$w_{\rm H}$ & Mr. Taniguchi said Denmark & 0 & 0 && 415 & 415 & 18 & -- & -- & F & $0$ & $0$ \\ \hline
	\end{tabular}
\end{table*}

\begin{table*}[!t]
\footnotesize
\caption{Itemized frequencies of unobserved four-grams for $S_{\rm tr}=2,500$.}
\renewcommand{\arraystretch}{1.3}
\label{tab:Itemized_2500}
\centering
	\begin{tabular}{ c  l  r  r  c  r  r  c  r  r  c  r  r } \hline
		\multicolumn{1}{ c }{\multirow{2}{*}{$w$}} & \multicolumn{1}{ c }{\multirow{2}{*}{$a_1a_2a_3a_4$}} & \multicolumn{2}{ c }{\#$\Psi_1w$} && \multicolumn{2}{ c }{\#$\Psi_2w$} && \multicolumn{2}{ c }{\#$\Psi_3w$} && \multicolumn{2}{ c }{\#$\Psi_4w$} \\ \cline{3-4} \cline{6-7} \cline{9-10} \cline{12-13}
		& & \multicolumn{1}{ c }{$O_{\rm ALL}$} & \multicolumn{1}{ c }{$O_{\rm LOC}$} && \multicolumn{1}{ c }{$O_{\rm ALL}$} & \multicolumn{1}{ c }{$O_{\rm LOC}$} &&  \multicolumn{1}{ c }{$O_{\rm ALL}$} & \multicolumn{1}{ c }{$O_{\rm LOC}$} && \multicolumn{1}{ c }{$O_{\rm ALL}$} & \multicolumn{1}{ c }{$O_{\rm LOC}$} \\ \hline \hline
		$w_{\rm C}$ & in West Germany, France, & 18,456 & 395 && 365 & 23 && 41 & 12 && 22 & 11 \\
		$w_{\rm D}$ & Christ Church in Grosse & 0 & 0 && 11 & 0 && 18,618 & 1,034 && 0 & 0 \\ \hline
		\multicolumn{1}{ c }{\multirow{2}{*}{$w$}} & \multicolumn{1}{ c }{\multirow{2}{*}{$a_1a_2a_3a_4$}} & \multicolumn{2}{ c }{\#$\Psi_1w$} && \multicolumn{2}{ c }{\#$\Psi_2w$} && \multicolumn{2}{ c }{\#$\Psi_3w$} && \multicolumn{2}{ c }{\#$\Psi_4w$} \\ \cline{3-4} \cline{6-7} \cline{9-10} \cline{12-13}
		& & \multicolumn{1}{ c }{$O_{\rm ALL}$} & \multicolumn{1}{ c }{$O_{\rm PER}$} && \multicolumn{1}{ c }{$O_{\rm ALL}$} & \multicolumn{1}{ c }{$O_{\rm PER}$} &&  \multicolumn{1}{ c }{$O_{\rm ALL}$} & \multicolumn{1}{ c }{$O_{\rm PER}$} && \multicolumn{1}{ c }{$O_{\rm ALL}$} & \multicolumn{1}{ c }{$O_{\rm PER}$} \\ \hline \hline
		$w_{\rm G}$ & promised by Prime Minister & 46 & 1 && 5,445 & 128 && 69 & 59 && 100 & 99 \\
		$w_{\rm H}$ & Mr. Taniguchi said Denmark & 5,704 & 141 && 0 & 0 && 6,170 & 161 && 3 & 0 \\ \hline
	\end{tabular}
\end{table*}

We analyzed the behavior of each method for $N=4$.
Table~\ref{tab:4-grams_2500} shows the word four-grams ranked by methods for $S_{\rm tr}=2,500$.
Table~\ref{tab:Itemized_2500} shows the itemized frequencies of the unobserved four-grams in Table~\ref{tab:4-grams_2500}.
$\Psi_kw$ for $w=a_1a_2 \cdots a_N$ represents $\bullet \bullet \cdots a_k \cdots \bullet$ as defined in Sect.~\ref{sec:Overall_Framework}.
The rank--recall curves corresponding to these tables are shown in Figs.~\ref{fig:Results_4-gram}(c) and \ref{fig:Results_4-gram}(d).
From the tables, we can see the following:

\begin{itemize}
  \item
  For B and K, all unobserved four-grams $w_{\rm C}$, $w_{\rm D}$, $w_{\rm G}$, and $w_{\rm H}$ were ranked at 611 and 415.
  This indicates that there were only 610 and 414 observed four-grams for LOC and PER, respectively, that could be LRs estimated by the two methods.
  Thus, B and K performed in a similarly poor manner.
  \item
  The behavior of KN was very different from that of the other approaches.
  The observed four-grams $w_{\rm A}$ and $w_{\rm E}$, which always occurred on the left side of the NEs in the training set, were not ranked in the top 8,000; instead, the higher ranks included the unobserved and false four-grams $w_{\rm D}$ and $w_{\rm H}$.
  Both of these four-grams are composed of unobserved units, as shown in Table~\ref{tab:Itemized_2500}, and are unreliable.
  This implies that taking the ratio of smoothed estimates resulted in poor performance.
  \item
  In focusing on unobserved four-grams, both ITEM and our method could ignore the false four-grams $w_{\rm D}$ and $w_{\rm H}$ and ranked the true four-grams $w_{\rm C}$ and $w_{\rm G}$ in high ranks.
  Nearly all units of $w_{\rm C}$ and $w_{\rm G}$ were frequent, as seen from Table~\ref{tab:Itemized_2500}.
  These results further confirm the effectiveness of itemizing $N$-grams.
  We could not confirm any difference in the results with and without $\widehat{t}_{\rm d}(w)$, as the chosen $\lambda_{\rm item}$ and $\lambda _{\rm d}$ were close values even in the LOC cases, as seen from Table~\ref{tab:Parameters_4-gram}.
  In addition, we confirmed many cases where $\widehat{t}_{\rm d}(w)$ was zero even in high ranks, as seen from $w_{\rm C}$ and $w_{\rm G}$.
  From the above, we can infer that $\widehat{t}_{\rm d}(w)$ does not work well in sparse data settings such as the 4-gram cases, and we leave the development of an effective approach to incorporating dependency to future work.
\end{itemize}

\subsection{Discussion}

In this section, we summarize the experimental results and discuss the benefits of our method and future works.

B was found to face both the low- and zero-frequency problems, and therefore its effectiveness is narrow and limited.
We confirmed that K handled low frequencies well but suffered from the zero-frequency problem.
We found that KN, which involves taking the ratio of smoothed estimates, was useless.
KN has eight parameters, which can conceivably be tuned simultaneously.
However, varying some probability estimators simultaneously no longer satisfies the smoothing scheme, which corrects a single estimator.

ITEM, involving the itemization of $N$-grams, was effective for rare and unobserved context prediction but sacrificed the dependency between items.
Thus, our method introduced the term $t_{\rm d}(w)$ to ITEM to incorporate the dependency.
This resulted in the best performance for two types of context prediction, which required different dependency strengths, suggesting that our method can control the dependency strength.
In experiments with the different $N$-gram order $N$s, the best comparison method was different for $N$, whereas our method consistently showed good performance.
In NLP, the proper $N$ sometimes varies depending on practical tasks.
Considering this fact, our method has the advantage that it can be used for various $N$s.
Furthermore, our method was found to be able to treat the entire high- to zero-frequency range within a unified framework that does not depend on a particular application.
This leads us to believe that our method will be helpful in various analyses and applications, including feature extraction, relation detection, and classification.

By contrast, $t_{\rm d}(w)$ measures dependency using only original $N$-gram frequencies and, for this reason, does not work well at sparse frequencies.
As an approach to incorporating dependency more effectively in future work, we could assume a multiple Markov independence in item occurrence.
We could also refer to previous research on alleviating the conditional independence assumption in naive Bayes classifiers~\cite{Frank:02,Jiang:12,Wu:16,Tang:16,Diab:17,Jiang:18} and consider whether these approaches are applicable to our method.
This might allow us to provide effective estimates from sparse data or sequences of long-$N$ $(\geq4)$ items.
As described in Sect.~\ref{sec:introduction}, likelihood ratios were used in numerous tasks.
Among these, our method can be used in cases where observation targets are continuous sequences (e.g., $N$-grams).
Another future work is to find further practical applications that use the sequences and verify the practicality of our method.

\section{Conclusion}

This paper presented a method for mitigating the low- and zero-frequency problems of LR estimation.
One possible approach for mitigating the zero-frequency problem is itemizing $w$ and roughly approximating $r(w)$ as $r_{\rm item}(w)$.
Although this is a straightforward approach, it ignores dependencies between items.
Therefore, our proposed method adds the term $t_{\rm d}(w)$, which is estimated from the $N$-gram frequencies, to $r_{\rm item}(w)$ to incorporate dependency and allow for the successful treatment of observed and unobserved $N$-grams.
To mitigate the low-frequency problem, we combined this framework with an existing method~\cite{Kikuchi:19}.
The two regularization parameters, $\lambda_{\rm item}$ and $\lambda_{\rm d}$, introduced by this combination, allowed us to control the dependency.
Our experimental results demonstrated that our method could mitigate both problems.
On the other hand, $t_{\rm d}(w)$ measured dependency using only the original $N$-gram frequencies.
Thus, we found that it did not work well when the frequency was sparse.
Hence, our future work includes the development of an approach to incorporate dependency more effectively.

\bibliography{references}
\bibliographystyle{ieicetr}

\appendix
\section{Derivation of the First Two Terms of the Objective Function in uLSIF}

In uLSIF, $\mbox{\boldmath $\beta$}$ are determined so that the squared loss $J_0(\mbox{\boldmath $\beta$})$ is minimized:
\begin{align*}
J_0(\mbox{\boldmath $\beta$}) =& \frac{1}{2} \int (\widehat{r}(x) - r(x))^2 p_{\rm de}(x) dx \\
=& \frac{1}{2} \int \widehat{r}(x)^2 p_{\rm de}(x) dx - \int \widehat{r}(x) r(x) p_{\rm de}(x) dx \\
& + \frac{1}{2} \int r(x)^2 p_{\rm de}(x) dx \\
=& \frac{1}{2} \int \widehat{r}(x)^2 p_{\rm de}(x) dx - \int \widehat{r}(x) p_{\rm nu}(x) dx \\
& + \frac{1}{2} \int r(x)^2 p_{\rm de}(x) dx,
\end{align*}
where $p_{\rm de}(x)$ is canceled in the second term.
In addition, the last term of $J_0(\mbox{\boldmath $\beta$})$ is a constant and therefore can be safely ignored.
Let the first two terms be $J(\mbox{\boldmath $\beta$})$:
\begin{align*}
J(\mbox{\boldmath $\beta$}) =& \frac{1}{2} \int \widehat{r}(x)^2 p_{\rm de}(x) dx - \int \widehat{r}(x) p_{\rm nu}(x) dx \\
=& \frac{1}{2} \sum_{l,l'=1}^b \beta_l \beta_{l'} \left( \int \varphi_l(x) \varphi_{l'}(x) p_{\rm de}(x) dx \right) \\
& - \sum_{l=1}^b \beta_l \left( \int \varphi_l(x) p_{\rm nu}(x) dx \right).
\end{align*}
Approximating the expectations in $J(\mbox{\boldmath $\beta$})$ by empirical averages, we obtain
\begin{align*}
\widehat{J}(\mbox{\boldmath $\beta$}) =& \frac{1}{2n_{\rm de}} \sum_{i=1}^{n_{\rm de}} \widehat{r}(x_i^{\rm de})^2 - \frac{1}{n_{\rm nu}} \sum_{j=1}^{n_{\rm nu}} \widehat{r}(x_j^{\rm nu}) \\
=& \frac{1}{2} \sum_{l,l'=1}^b \beta_l \beta_{l'} \left( \frac{1}{n_{\rm de}} \sum_{i=1}^{n_{\rm de}} \varphi_l(x_i^{\rm de}) \varphi_{l'}(x_i^{\rm de}) \right) \\
& - \sum_{l=1}^b \beta_l \left( \frac{1}{n_{\rm nu}} \sum_{j=1}^{n_{\rm nu}} \varphi_l(x_j^{\rm nu}) \right) \\
=& \frac{1}{2} \mbox{\boldmath $\beta$}^{\mathrm{T}} \widehat{\mbox{\boldmath $H$}} \mbox{\boldmath $\beta$} - \widehat{\mbox{\boldmath $h$}}^{\mathrm{T}} \mbox{\boldmath $\beta$},
\end{align*}
which are the first two terms of the objective function in Eq.(3).

\section{Derivation of $\widehat{J}(\mbox{\boldmath $\beta$})$ in Our Method}

Let the squared loss be $J_0(\mbox{\boldmath $\beta$})$:
\begin{align*}
J_0(\mbox{\boldmath $\beta$}) =& \frac{1}{2} \int (r_{\rm ours}(w) - r(w))^2 p_{\rm de}(w) dw \\
=& \frac{1}{2} \int r_{\rm ours}(w)^2 p_{\rm de}(w) dw \\
& - \int r_{\rm ours}(w) r(w) p_{\rm de}(w) dw \\
& + \frac{1}{2} \int r(w)^2 p_{\rm de}(w) dw \\
=& \frac{1}{2} \int r_{\rm ours}(w)^2 p_{\rm de}(w) dx \\
& - \int r_{\rm ours}(w) p_{\rm nu}(w) dw \\
& + \frac{1}{2} \int r(w)^2 p_{\rm de}(w) dw,
\end{align*}
where $p_{\rm de}(w)$ is canceled in the second term and the last term can be safely ignored.
Then, let the first two terms be $J(\mbox{\boldmath $\beta$})$:
\begin{align*}
J(\mbox{\boldmath $\beta$}) =& \frac{1}{2} \int r_{\rm ours}(w)^2 p_{\rm de}(w) dw \\
& - \int r_{\rm ours}(w) p_{\rm nu}(w) dw \\
=& \frac{1}{2} \sum_{l,l'=1}^v \beta_l \beta_{l'} \left( \int \varphi_l(w) \varphi_{l'}(w) p_{\rm de}(w) dw \right) \\
& + \widehat{r}_{\rm item}(w) \sum_{l=1}^v \beta_l \left( \int \varphi_l(w) p_{\rm de}(w) dw \right) \\
& - \sum_{l=1}^v \beta_l \left( \int \varphi_l(w) p_{\rm nu}(w) dw \right) \\
& + \frac{1}{2} \widehat{r}_{\rm item}(w)^2 \int p_{\rm de}(w) dw \\
& - \widehat{r}_{\rm item}(w) \int p_{\rm nu}(w) dw,
\end{align*}
where $r_{\rm item}(w)$ is replaced by $\widehat{r}_{\rm item}(w)$, as shown in Eq.(10), and treated as a constant.
Thus, the last two terms are also constants and can be safely ignored.
Approximating the expectations of the first three terms in $J(\mbox{\boldmath $\beta$})$ by empirical averages, we obtain
\begin{align*}
\widehat{J}(\mbox{\boldmath $\beta$}) =& \frac{1}{2}\sum_{l,l'=1}^v \beta_{l} \beta_{l'} \left( \frac{1}{n_{\rm de}} \sum_{i=1}^{n_{\rm de}} \varphi_l(w_i^{\rm de}) \varphi_{l'}(w_i^{\rm de}) \right) \\
& + \widehat{r}_{\rm item}(w) \sum_{l=1}^v \beta_l \left( \frac{1}{n_{\rm de}} \sum_{i=1}^{n_{\rm de}} \varphi_l(w_i^{\rm de}) \right) \\
& - \sum_{l=1}^v \beta_l \left( \frac{1}{n_{\rm nu}} \sum_{j=1}^{n_{\rm nu}} \varphi_l(w_j^{\rm nu}) \right),
\end{align*}
where the first term becomes zero unless $l = l'$ and therefore can be rewritten as
\begin{align*}
\widehat{J}(\mbox{\boldmath $\beta$}) =& \frac{1}{2}\sum_{l=1}^v {\beta_l}^2 \left( \frac{1}{n_{\rm de}} \sum_{l=1}^{n_{\rm de}} \varphi_l(w_i^{\rm de}) \right) \\
& + \widehat{r}_{\rm item}(w) \sum_{l=1}^v \beta_l \left( \frac{1}{n_{\rm de}} \sum_{i=1}^{n_{\rm de}} \varphi_l(w_i^{\rm de}) \right) \\
& - \sum_{l=1}^v \beta_l \left( \frac{1}{n_{\rm nu}} \sum_{j=1}^{n_{\rm nu}} \varphi_l(w_j^{\rm nu}) \right).
\end{align*}
From the above, we obtain $\widehat{J}(\mbox{\boldmath $\beta$})$ as in Eq.(13).

\section{Precision--Recall Curves}

\begin{figure}[!t]
  \begin{minipage}[b]{0.5\linewidth}
    \centering
    \includegraphics[keepaspectratio, scale=0.28]{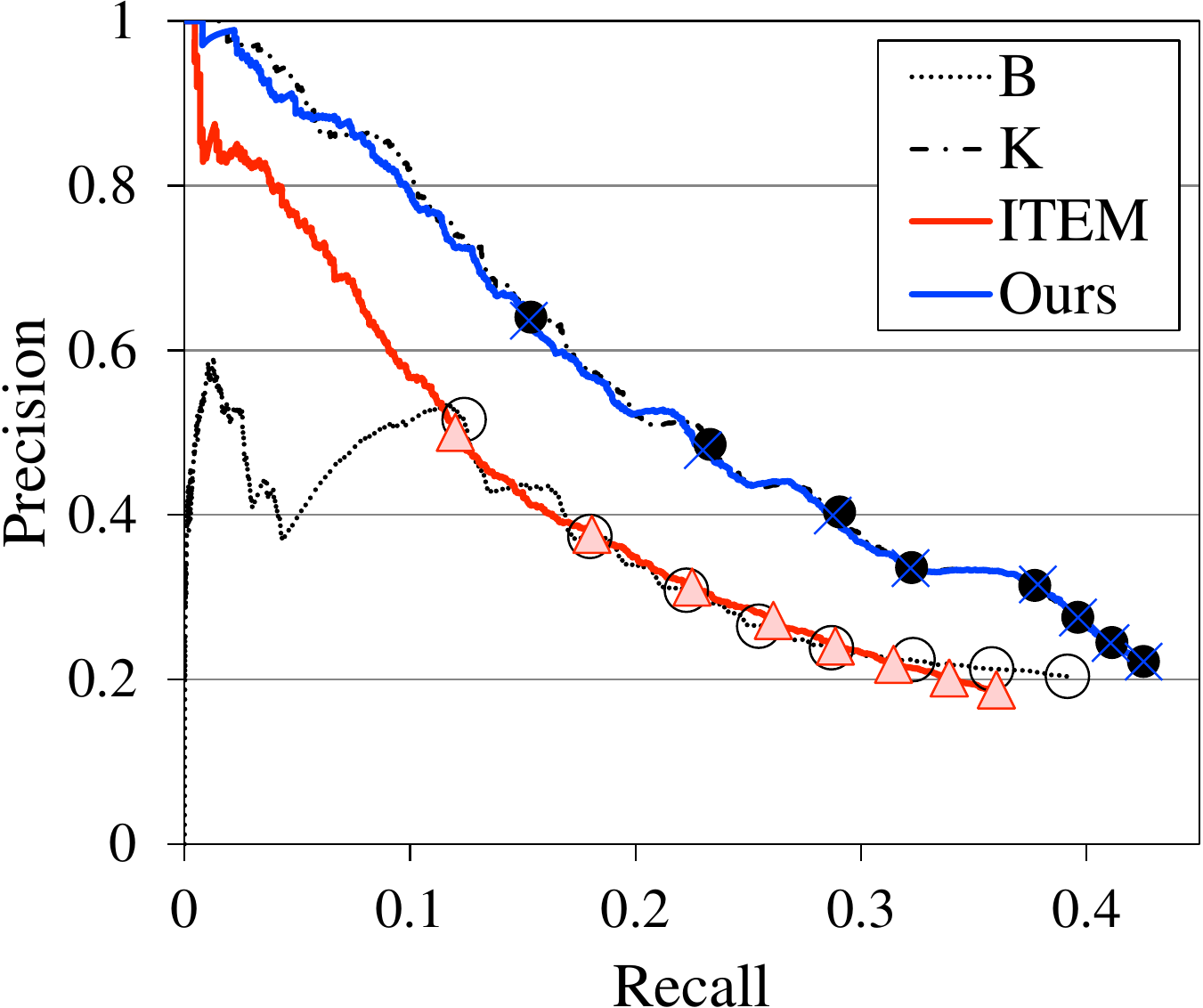}
    \subcaption{LOC, $S_{\rm tr}=10,000$}\label{fig:PR_LOC_10000_2-gram}
  \end{minipage}
  \begin{minipage}[b]{0.5\linewidth}
    \centering
    \includegraphics[keepaspectratio, scale=0.28]{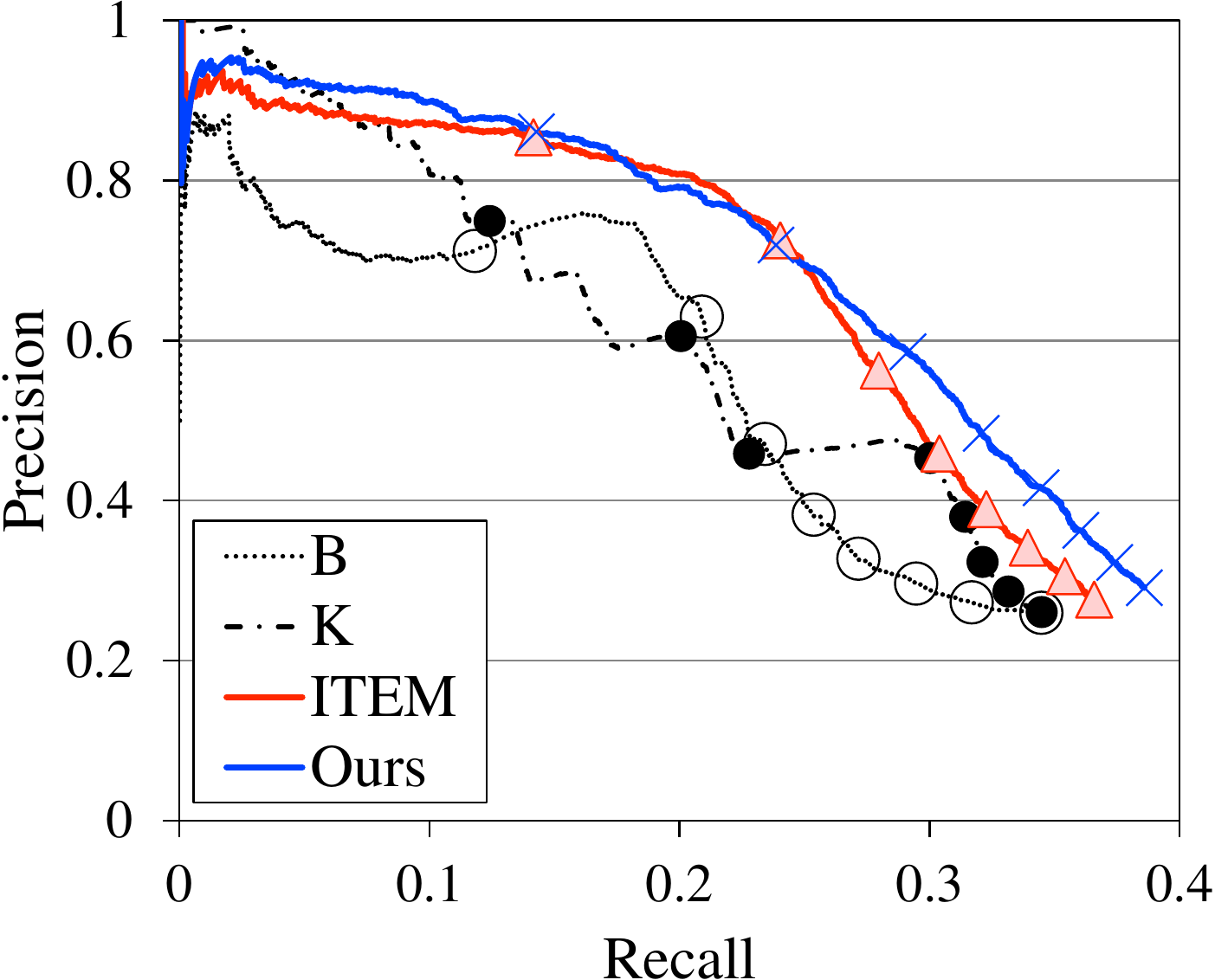}
    \subcaption{PER, $S_{\rm tr}=10,000$}\label{fig:PR_PER_10000_2-gram}
  \end{minipage} \vspace{1pt} \\
  \begin{minipage}[b]{0.5\linewidth}
    \centering
    \includegraphics[keepaspectratio, scale=0.28]{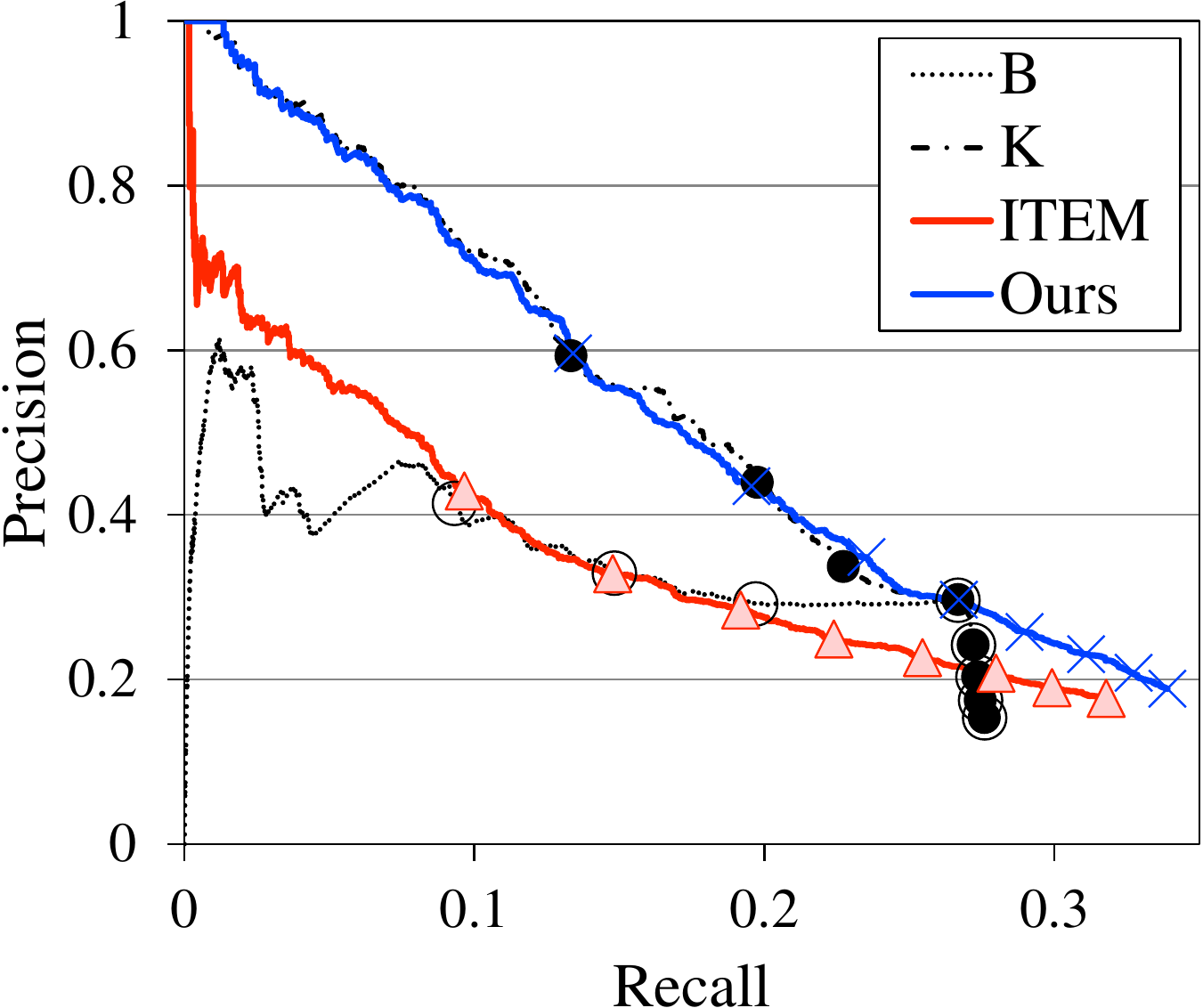}
    \subcaption{LOC, $S_{\rm tr}=2,500$}\label{fig:PR_LOC_2500_2-gram}
  \end{minipage}
  \begin{minipage}[b]{0.5\linewidth}
    \centering
    \includegraphics[keepaspectratio, scale=0.28]{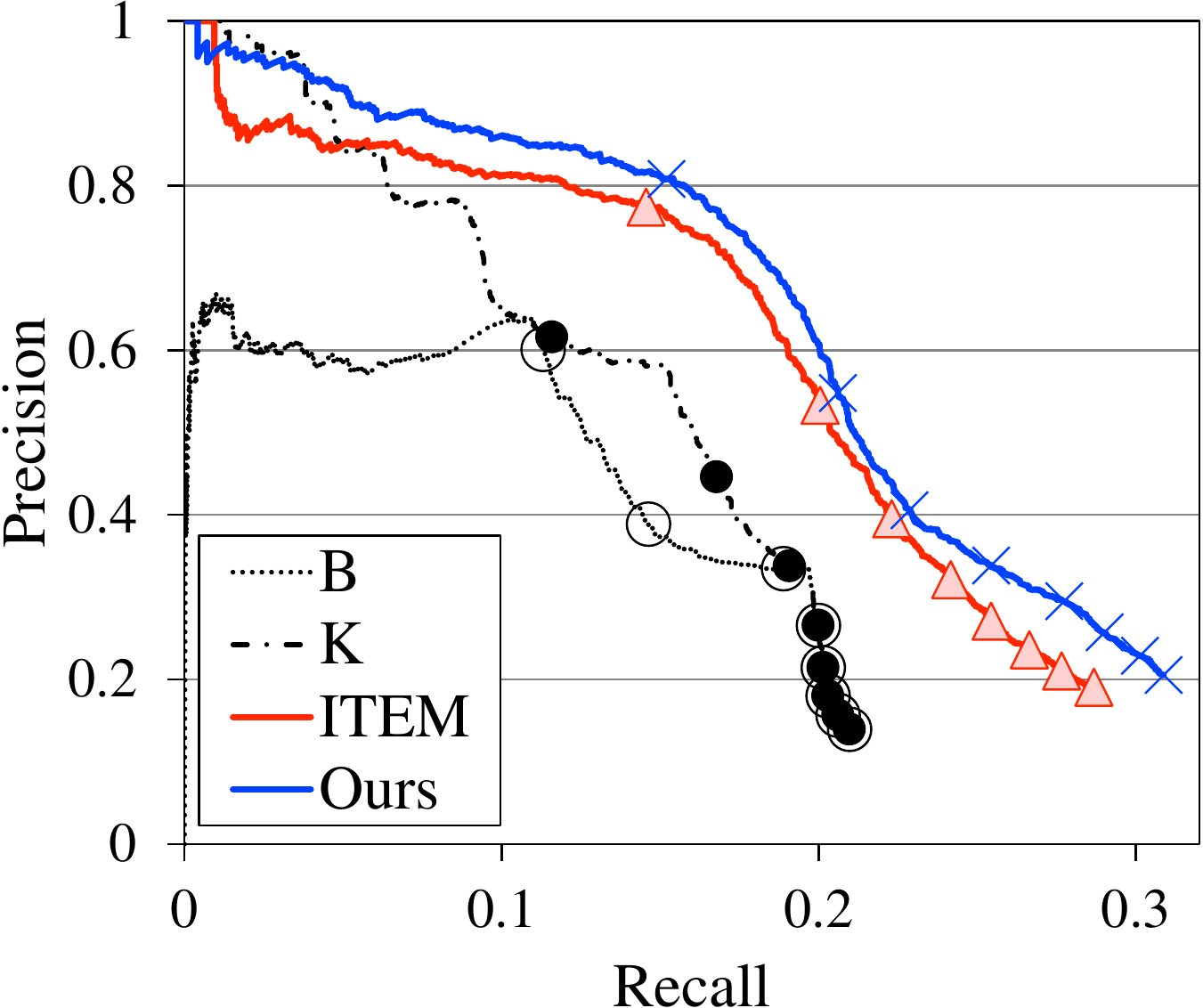}
    \subcaption{PER, $S_{\rm tr}=2,500$}\label{fig:PR_PER_2500_2-gram}
  \end{minipage}
  \caption{
  Precision--recall curves for $N=2$.
  PER and LOC indicate that prediction targets are the left PER and LOC contexts, respectively.
  $S_{\rm tr}$ is the training set size.
  The horizontal and vertical axes of curves represent recall and precision, respectively.
  The markers on each curve indicate points of ranks 1,000, 2,000, $\ldots$, 8,000.
  For a given rank, the method with the largest values on the horizontal and vertical axes, i.e., the highest recall and precision, has the best performance.
  }
  \label{fig:PR_2-gram}
\end{figure}

\begin{figure}[!t]
  \begin{minipage}[b]{0.5\linewidth}
    \centering
    \includegraphics[keepaspectratio, scale=0.28]{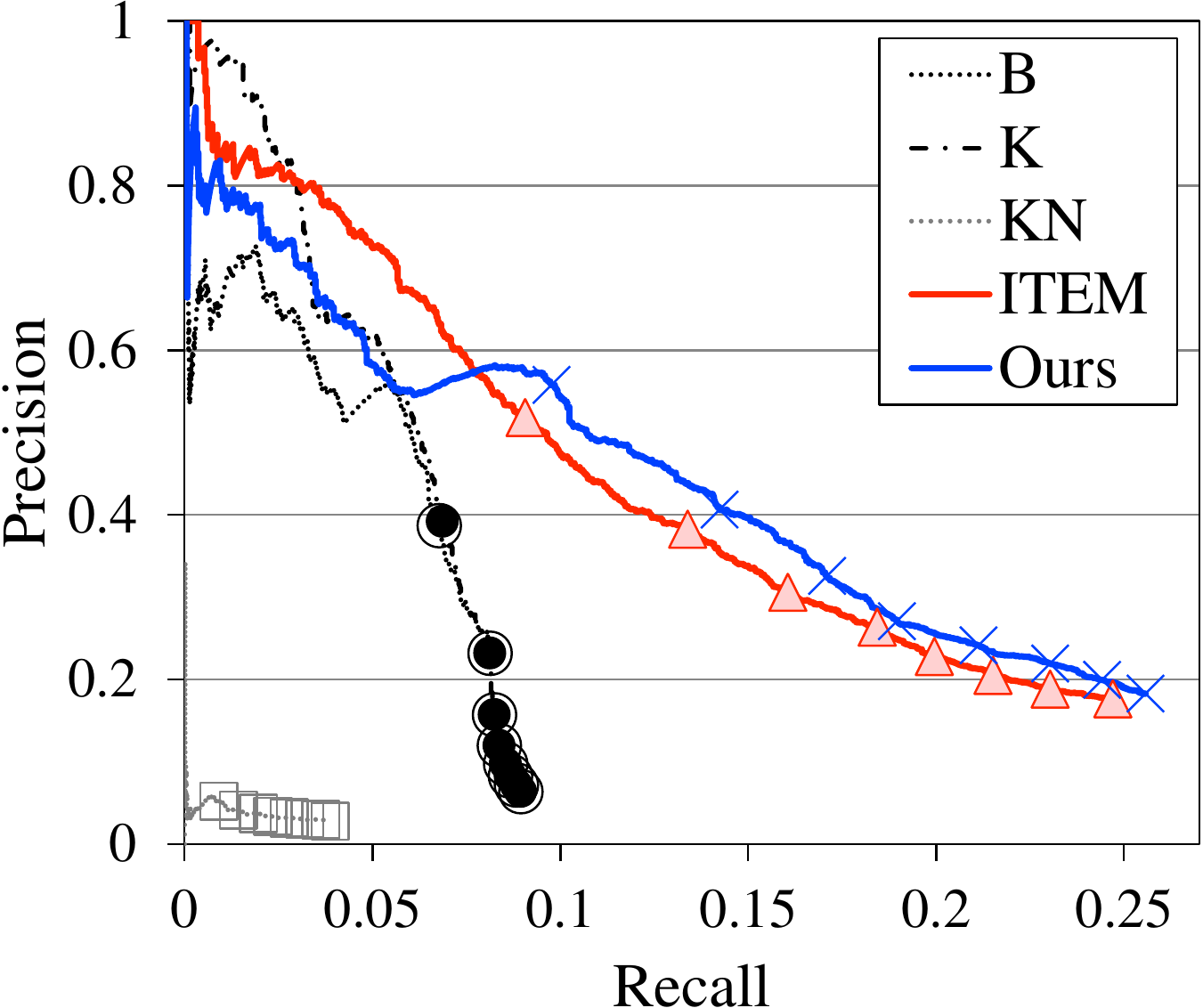}
    \subcaption{LOC, $S_{\rm tr}=10,000$}\label{fig:PR_LOC_10000_4-gram}
  \end{minipage}
  \begin{minipage}[b]{0.5\linewidth}
    \centering
    \includegraphics[keepaspectratio, scale=0.28]{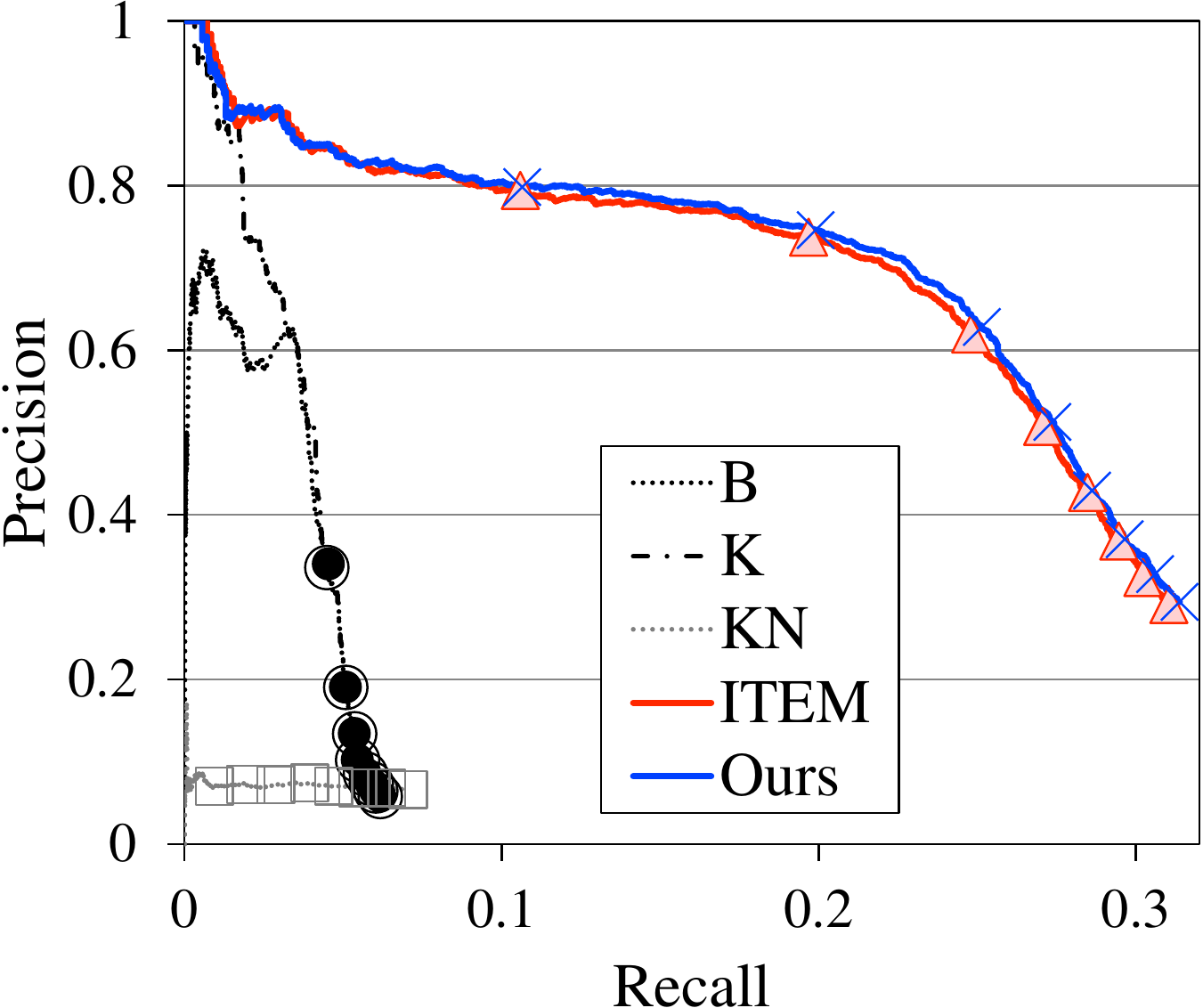}
    \subcaption{PER, $S_{\rm tr}=10,000$}\label{fig:PR_PER_10000_4-gram}
  \end{minipage} \vspace{1pt} \\
  \begin{minipage}[b]{0.5\linewidth}
    \centering
    \includegraphics[keepaspectratio, scale=0.28]{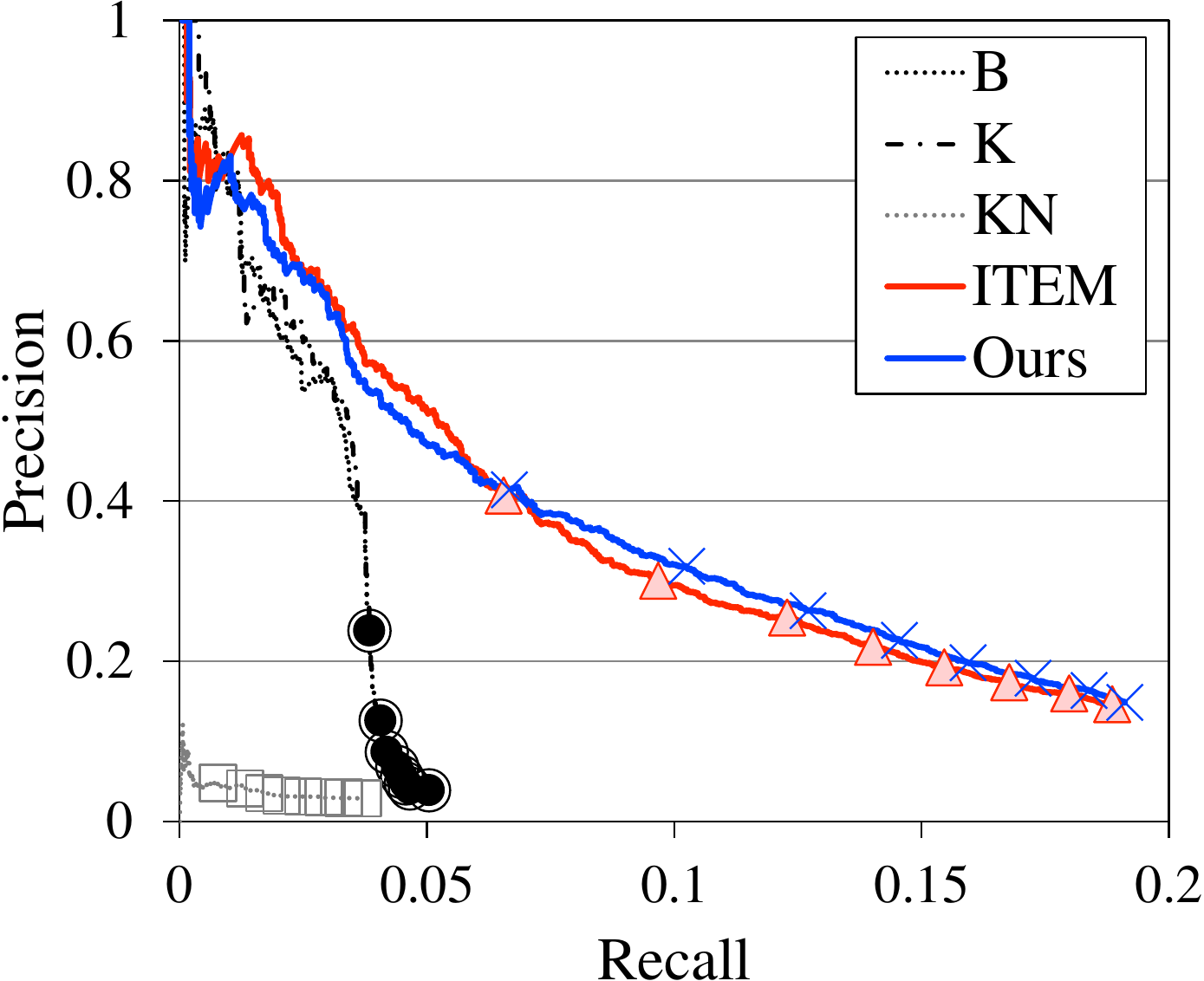}
    \subcaption{LOC, $S_{\rm tr}=2,500$}\label{fig:PR_LOC_2500_4-gram}
  \end{minipage}
  \begin{minipage}[b]{0.5\linewidth}
    \centering
    \includegraphics[keepaspectratio, scale=0.28]{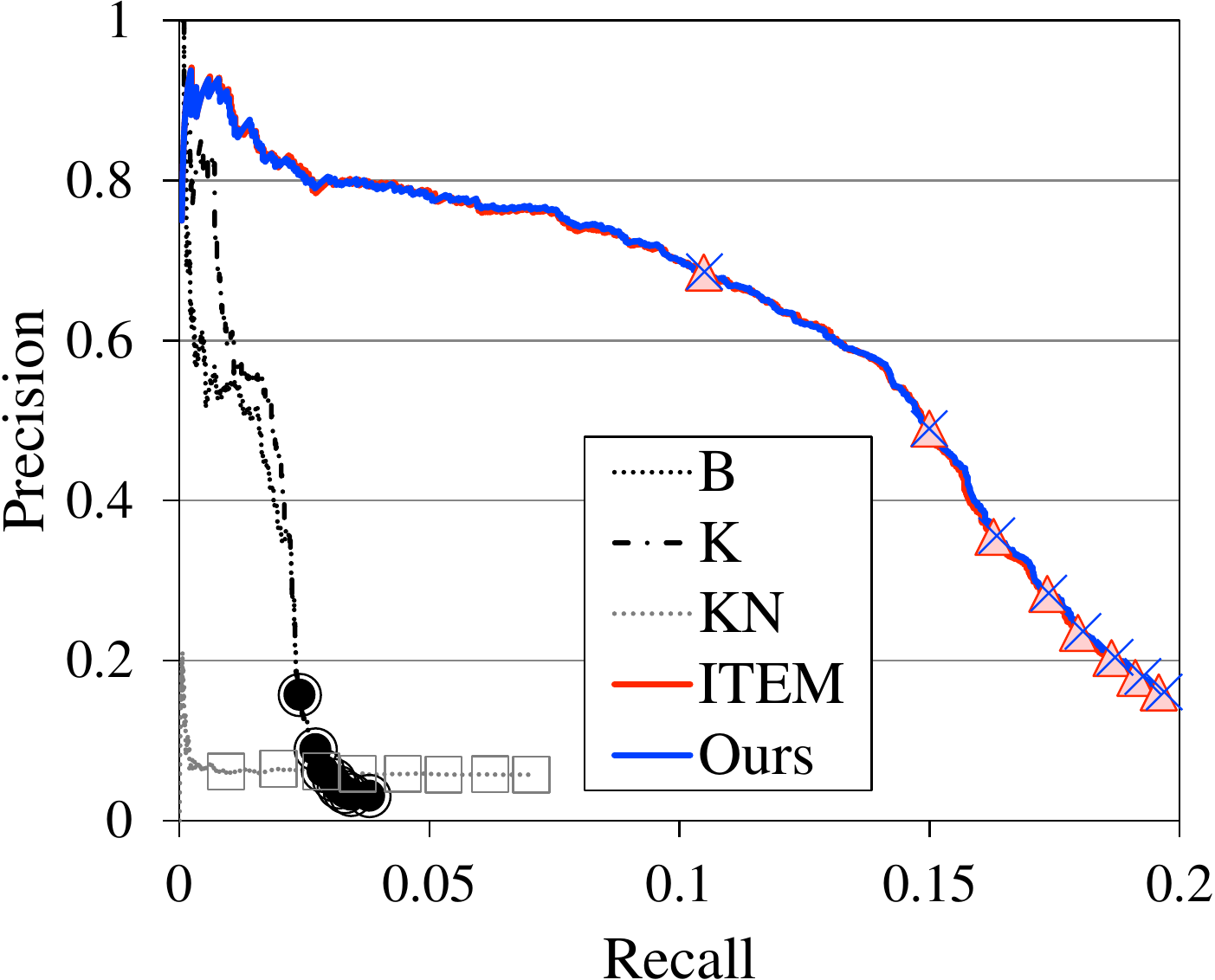}
    \subcaption{PER, $S_{\rm tr}=2,500$}\label{fig:PR_PER_2500_4-gram}
  \end{minipage}
  \caption{
  Precision--recall curves for $N=4$.
  PER and LOC indicate that prediction targets are the left PER and LOC contexts, respectively.
  $S_{\rm tr}$ is the training set size.
  The horizontal and vertical axes of curves represent recall and precision, respectively.
  The markers on each curve indicate points of ranks 1,000, 2,000, $\ldots$, 8,000.
  For a given rank, the method with the largest values on the horizontal and vertical axes, i.e., the highest recall and precision, has the best performance.
  }
  \label{fig:PR_4-gram}
\end{figure}

Figs.~\ref{fig:PR_2-gram} and \ref{fig:PR_4-gram} show precision--recall curves representing different forms of the rank--recall curves shown in Figs. 5 and 6, respectively.
In Sect. 5, the parameters for each method were chosen based on the top 8,000 $N$-grams ranked in descending order of estimates.
Therefore, the curves shown here are also depicted using only the top 8,000.

\profile{Masato KIKUCHI}{
received the B.E., M.E. and Ph.D. degrees from Toyohashi University of Technology, Japan, in 2016, 2018 and 2021, respectively. He is currently an assistant professor of the Department of Computer Science, Graduate School of Engineering, Nagoya Institute of Technology. His research interests include data mining, natural language processing, and web intelligence.
}
\profile{Kento KAWAKAMI}{
received the B.E. and M.E. degrees from Toyohashi University of Technology, Japan, in 2017 and 2019, respectively.
His research interests include data mining, natural language processing, and network security.
}
\profile{Kazuho WATANABE}{
received the B.E., M.E. and Ph.D. degrees from Tokyo Institute of Technology, Japan, in 2002, 2004 and 2006, respectively.
He is currently an associate professor of the Department of Computer Science and Engineering, Toyohashi University of Technology.
His research interests include statistical machine learning theory and algorithms.
}
\profile{Mitsuo YOSHIDA}{
received the Ph.D. degree from University of Tsukuba, Japan, in 2014.
He is currently an assistant professor of the Department of Computer Science and Engineering, Toyohashi University of Technology.
He is also the founder of TechTech Inc., which provides a news search engine and others. 
His research interests include the science of science, computational social science, and natural language processing.
}
\profile[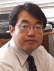]{Kyoji UMEMURA}{
received the B.E., M.E., and Ph.D. degrees from the University of Tokyo, Japan, in 1981, 1983 and 1991, respectively.
He is currently a professor of the Department of Computer Science and Engineering, Toyohashi University of Technology.
His research interests include information retrieval, Lisp and symbolic computation, compiler, operating system, and natural language processing.
}

\end{document}